\documentclass{article}
\usepackage{iclr2027_conference,times}
\setcitestyle{authoryear,round,semicolon}
\usepackage{hyperref,url}
\usepackage{amsmath,amssymb,amsfonts,amsthm,mathtools}
\usepackage{graphicx,booktabs,array,multirow,enumitem,xcolor,microtype,xspace}
\usepackage{subcaption,adjustbox,makecell,colortbl}
\usepackage{algorithm,algorithmic}
\hypersetup{hidelinks}
\theoremstyle{plain}
\newtheorem{proposition}{Proposition}
\newtheorem{corollary}{Corollary}
\newtheorem{theorem}{Theorem}
\title{Persistent Tri-State Message Passing}

\author{Yoonhyuk Choi \\
Sookmyung Women's University \\
\texttt{chldbsgur123@gmail.com}
\And
Jiho Choi \\
KAIST \\
\texttt{jihochoi1993@gmail.com}
\And
Chong-Kwon Kim \\
Korea Institute of Energy Technology \\
\texttt{ckim@kentech.ac.kr}
}

\iclrfinalcopy

\date{}
\begin{document}
\maketitle

\lhead{Preprint}

\begin{abstract}
In stochastic message passing, an edge's sampled role changes the node states used to compute adaptive weights at later layers. Weight averaging therefore depends on whether edge roles persist across layers or are resampled at each layer. We study this dependence in Persistent Tri-State Message Passing (P3MP), which combines persistent additive, subtractive, and inactive roles with weights computed from each sample's endpoint states. For two layers, we separate target-state feedback from weight-source covariance and derive the condition under which local and shared weights reverse their preactivation ordering. Shared weights retain a $1/K$ fraction of the feedback for $K$ samples. The interaction is zero for a source-only scorer. Permuting complete weight vectors between samples quantifies the same role-state dependence while preserving their empirical distribution. Exact enumeration and feature-only checkpoint measurements agree with the analytical predictions. Across independently trained feature-only models, local weights yield higher accuracy and lower negative log-likelihood under persistence, while shared weights yield higher accuracy and lower loss under layerwise resampling. The same accuracy ordering holds on deduplicated Chameleon and Squirrel. 
\end{abstract}

\section{Introduction}
Graph neural networks combine node features with messages from their neighbors \citep{gilmer2017neural,hamilton2017inductive}. A neighbor's contribution depends on the prediction task, especially when connected nodes have different labels. Its message can be added, subtracted, or set to zero \citep{zhu2020beyond,bo2021beyond,chien2021adaptive,choi2025beyond,choi2025selective}. When weights depend on intermediate node representations, message operations affect later layers by changing the states used to compute subsequent weights.

Consider an edge whose role is sampled once and used throughout the layers. Its first message changes the target representation, which the next scorer uses to compute the edge's weight. The later gate is paired with a state that depends on the edge's earlier role. Under layerwise resampling, the next gate is independent of the earlier draw. With identical marginal role probabilities, the two sampling rules can still produce different dependencies between gates and adaptive weights. Cross-sample weight averaging changes how these dependencies contribute to the propagated messages.

DropEdge studies both perturbed adjacencies shared across layers and adjacencies sampled independently at each layer \citep{rong2019dropedge}. Bayesian graph models average predictions over sampled structures, and adaptive connection sampling learns structural probabilities jointly with the predictor \citep{zhang2019bayesian,hasanzadeh2020bayesian}. We analyze the effect of adaptive weight averaging under persistent and resampled roles. In a two-layer computation, we separate feedback through the target state from covariance between a weight and its source value. The relative signs and magnitudes of these terms determine when local and shared weights reverse their preactivation ordering between the two sampling rules.

Persistent Tri-State Message Passing (P3MP) uses additive, subtractive, and inactive edge roles with thresholded endpoint weights and is trained with mixture negative log-likelihood. Each sampled role persists across layers. At every layer, the weights are recomputed from that sample's current states. The feedback decomposition applies to binary and other finite-valued gates as well as the additive, subtractive, and inactive operations in P3MP.

In independently trained feature-only models, local weights yield higher accuracy under persistence, and shared weights yield higher accuracy under resampling. Across the five graphs in Table \ref{tab:feature_factorial}, the accuracy benefit of local over shared weighting is 1.1 to 2.5 percentage points larger under persistence. This ordering also holds after duplicate removal on Chameleon and Squirrel. At feature-only checkpoints, the endpoint-scorer interaction is positive, and the source-only response is close to zero, consistent with the analysis. We examine role distributions and predictive averaging in a label-aware extension.

\section{Preliminaries}
Let $\mathcal G_{\mathrm{obs}}=(\mathcal V,\mathcal E_{\mathrm{obs}})$ have $n$ nodes and $m$ directed edges. We use target-source order: $(i,j)$ carries a message from $j$ to $i$, and $A_{\mathrm{obs},ij}=1$ indicates its presence. The feature matrix is $X\in\mathbb R^{n\times d}$, where $d$ is the feature dimension, and labels $y_i\in\{1,\ldots,C\}$ range over $C$ classes. We use the disjoint sets $\mathcal V_{\mathrm{tr}}$, $\mathcal V_{\mathrm{val}}$, and $\mathcal V_{\mathrm{te}}$ for supervision, model selection, and evaluation. Transductive propagation uses all observed features and edges. We write $h_i^{(\ell)}$ for node $i$'s layer-$\ell$ representation and $\mathcal N_i(A_{\mathrm{obs}})$ for its incoming neighborhood. The structural input is $X_{\mathrm{st}}=X$ in the feature-only model. The label-aware extension is defined in Appendix \ref{sec:label_extension}. We summarize the graph's label agreement by global edge homophily \citep{zhu2020beyond}:
\begin{equation}
\mathcal G_h=\frac1m\sum_{(i,j)\in\mathcal E_{\mathrm{obs}}}\mathbb I(y_i=y_j).
\label{global_homophily}
\end{equation}

\section{Methodology}\label{sec:method}
\begin{figure*}[t]
\centering
\includegraphics[width=\textwidth]{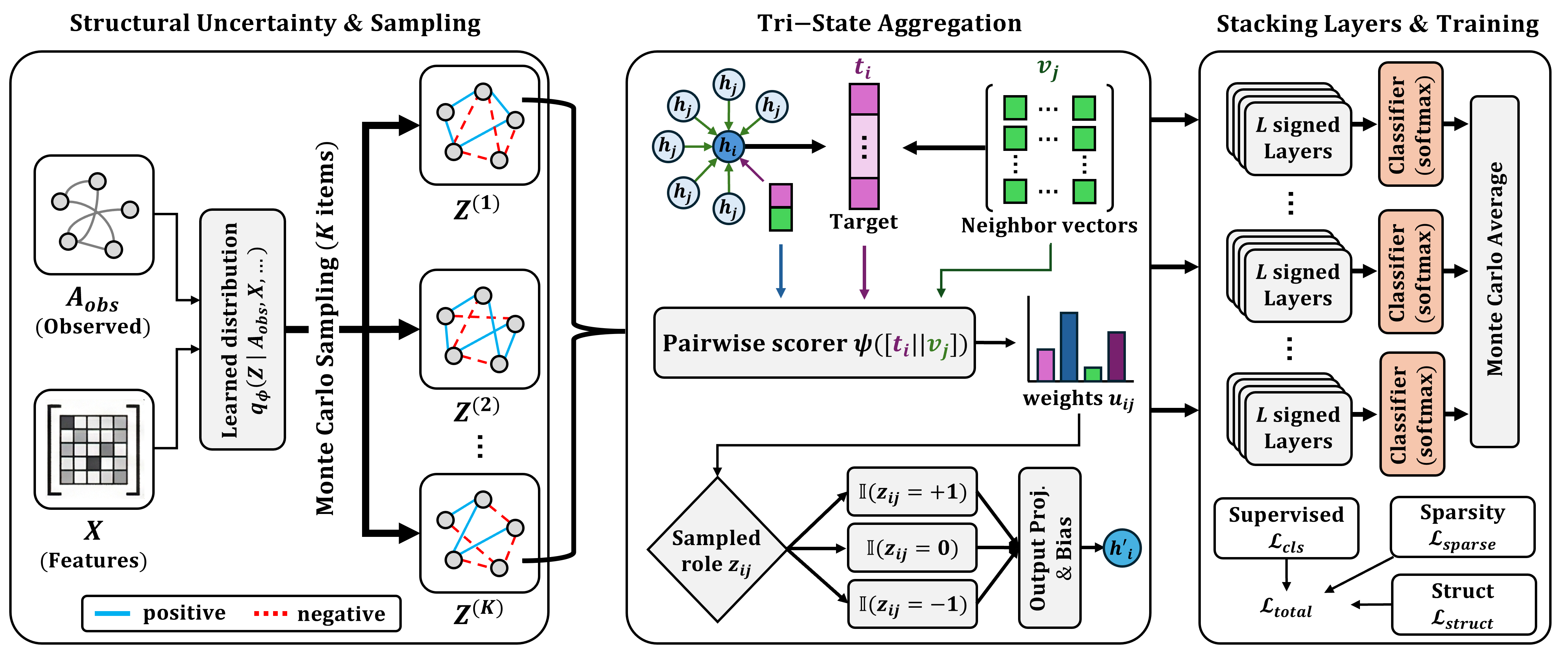}
\caption{P3MP's forward computation. The structural network assigns probabilities to subtractive, inactive, and additive roles, which persist through the signed layers. Endpoint scorers recompute weights $u_{ij}$, and a shared classifier supplies the probabilities for predictive averaging. Training combines classification, sparsity, and structural losses. Equation \ref{eq:signed-aggregation} includes the learned self-contribution.}
\label{model}
\end{figure*}

P3MP combines prediction and structural networks (Figure \ref{model}). Prediction parameters $\theta$ comprise the backbone, signed layers, endpoint scorers, and classifier. Structural parameters $\phi$ belong to a graph encoder and edge-role decoder, which define the distribution of roles used by the prediction network.

\subsection{Persistent edge roles}
We initialize the prediction network with $H^{(0)}=B_\theta(X,A_{\mathrm{obs}})$, using a backbone of hidden width 128. A separate encoder computes $H_\phi=\mathrm{GNN}_\phi(A_{\mathrm{obs}},X_{\mathrm{st}})$, with node representation $h_{\phi,i}$ in row $i$. An edge decoder $D_\phi$ gives the role probabilities:
\begin{equation}
\pi_{ij}=\operatorname{softmax}\bigl(D_\phi(h_{\phi,i},h_{\phi,j})\bigr),\qquad
q_\phi(Z\mid A_{\mathrm{obs}},X_{\mathrm{st}})
=\prod_{(i,j)\in\mathcal E_{\mathrm{obs}}}\operatorname{Cat}(z_{ij};\pi_{ij}).
\end{equation}
The entries of $\pi_{ij}=(\pi^-_{ij},\pi^0_{ij},\pi^+_{ij})$ correspond to subtractive, inactive, and additive roles, and $\operatorname{Cat}$ denotes the categorical law. A draw $z_{ij}\in\{-1,0,+1\}$ selects an edge's operation, with $Z$ collecting all edge roles. Endpoint embeddings provide the decoder with graph context. We sample roles independently given these embeddings, and set entries outside the observed support to zero.

For sample $k$, we draw $Z^{(k)}\sim q_\phi$ once and use it through $L$ signed layers, starting from $H^{(0,k)}=H^{(0)}$. All samples share learned parameters. Training uses hard straight-through Gumbel-softmax sampling at backward temperature $T=0.5$ \citep{jang2016categorical}, and prediction uses categorical roles with dropout disabled.

\subsection{Adaptive edge weights}
At each signed layer, we compute an edge's weight from its current endpoint representations. With layer and sample indices suppressed, we use
\begin{equation}
t_i=W_t h_i,\qquad v_j=W_v h_j,\qquad
s_{ij}=\psi_\theta(t_i,v_j),\qquad
u_{ij}=[|s_{ij}|-\lambda_{\mathrm{sc}}]_+.
\label{eq:mass}
\end{equation}
The learned maps $W_t$ and $W_v$ produce the width-64 target projection $t_i$ and source value $v_j$. The scorer $\psi_\theta$ concatenates them and applies a two-layer perceptron with hidden width 128 and scalar output $s_{ij}$. The positive-part operator is $[x]_+=\max\{x,0\}$. With threshold $\lambda_{\mathrm{sc}}=0.05$, the weights $u_{ij}$ are nonnegative and unnormalized across neighbors.

The sampled role determines a signed gate $d_{ij}(Z)$ and an activity indicator $b_{ij}(Z)$:
\begin{equation}
d_{ij}(Z)=\mathbb I(z_{ij}=+1)-\gamma\mathbb I(z_{ij}=-1),\qquad
b_{ij}(Z)=\mathbb I(z_{ij}\ne0).
\end{equation}
We learn a separate subtractive strength $\gamma>0$ in each layer through a softplus parameterization. The signed update is
\begin{equation}
h_i'=W_o\sum_{j\in\mathcal N_i(A_{\mathrm{obs}})}u_{ij}d_{ij}(Z)v_j
+W_{\mathrm{self}}h_i+b.
\label{eq:signed-aggregation}
\end{equation}
The output map $W_o\in\mathbb R^{128\times64}$ maps the aggregate to hidden space. We add a learned self-contribution through $W_{\mathrm{self}}\in\mathbb R^{128\times128}$ and bias $b\in\mathbb R^{128}$. An inactive role produces a zero message throughout the sample. For an active role, transmission also depends on whether the layer's score exceeds the threshold, so its message coefficient is zero when $|s_{ij}|\le\lambda_{\mathrm{sc}}$. Message sparsity describes this exact support.

Applying the signed update at every node gives
\begin{equation}
H^{(\ell+1,k)}=\operatorname{ReLU}\bigl(A^{(\ell+1,k)}\bigr),
\qquad \ell=0,\ldots,L-1.
\end{equation}
The rows of $A^{(\ell+1,k)}$ are the preactivations in Eq. \ref{eq:signed-aggregation}, evaluated from $H^{(\ell,k)}$ and $Z^{(k)}$. Each layer has its own parameters. During training, all samples share the backbone dropout realization, while dropout on the signed activations is independent. The next scorer then uses the corresponding sample's perturbed states. Scores and gates are evaluated on every observed edge, preserving the straight-through derivatives associated with inactive draws (Appendix \ref{sec:app:execution}).

\subsection{Learning the predictive average}
We apply a shared linear classifier to obtain the sample probabilities $p_i^{(k)}$, then average them over $K$ samples to form $\hat p_i^{(K)}$:
\begin{equation}
p_i^{(k)}=\operatorname{softmax}(W_c h_i^{(L,k)}+c),\qquad
\hat p_i^{(K)}=\frac1K\sum_{k=1}^Kp_i^{(k)}.
\end{equation}
The classifier parameters are $W_c\in\mathbb R^{C\times128}$ and $c\in\mathbb R^C$. Let $P_{ik}=p_{i,y_i}^{(k)}$ denote the target-class probability in sample $k$. We train the average with mixture NLL, which gives each sample a responsibility $r_{ik}$:
\begin{equation}
\mathcal L_{\mathrm{cls},i}=-\log\left(\frac1K\sum_kP_{ik}\right),\qquad
r_{ik}=\frac{P_{ik}}{\sum_bP_{ib}},\qquad
\frac{\partial\mathcal L_{\mathrm{cls},i}}{\partial\log P_{ik}}=-r_{ik}.
\label{eq:responsibilities}
\end{equation}
The responsibility $r_{ik}$ is sample $k$'s share of the target probability for node $i$. Larger contributions receive larger gradient weights, while the mean sample NLL assigns weight $1/K$ to every sample.

We regularize active edge weights through
\begin{equation}
\mathcal L_{\mathrm{sparse}}=
\frac1{nLK}\sum_{k=1}^K\sum_{\ell=0}^{L-1}
\sum_{(i,j)\in\mathcal E_{\mathrm{obs}}}u_{ij}^{(\ell,k)}b_{ij}(Z^{(k)}).
\end{equation}
We use the same straight-through estimator for the activity indicators and signed gates. The structural regularizer combines a Kullback-Leibler (KL) penalty against uniform role probabilities with an activity term:
\begin{equation}
\begin{aligned}
\mathcal L_{\mathrm{KL}}&=\frac1m\sum_{(i,j)\in\mathcal E_{\mathrm{obs}}}
\sum_{s\in\{-,0,+\}}\pi_{ij}^s\log(3\pi_{ij}^s),\\
\mathcal L_{\mathrm{act}}&=-\frac1m\sum_{(i,j)\in\mathcal E_{\mathrm{obs}}}
\log(\pi_{ij}^++\pi_{ij}^-),\\
\mathcal L_{\mathrm{struct}}&=\lambda_{\mathrm{KL}}\mathcal L_{\mathrm{KL}}+
\lambda_{\mathrm{act}}\mathcal L_{\mathrm{act}}.
\end{aligned}
\label{eq:struct-loss}
\end{equation}
We set $\lambda_{\mathrm{KL}}=\lambda_{\mathrm{act}}=1$. The structural terms encourage diffuse role probabilities and activity, with equal preference for the additive and subtractive states (Appendix \ref{sec:app:regularizer}). The weight penalty controls the magnitudes of active weights. We optimize the complete objective
\begin{equation}
\mathcal L_{\mathrm{total}}=
\frac1{|\mathcal V_{\mathrm{tr}}|}\sum_{i\in\mathcal V_{\mathrm{tr}}}\mathcal L_{\mathrm{cls},i}
+\lambda_{\mathrm{sp}}\mathcal L_{\mathrm{sparse}}
+\lambda_{\mathrm{st}}\mathcal L_{\mathrm{struct}}.
\label{eq:total}
\end{equation}
We set the sparsity and structural loss weights to $\lambda_{\mathrm{sp}}=0.01$ and $\lambda_{\mathrm{st}}=0.1$, with five training and prediction samples and default signed depth $L=2$. We update $(\theta,\phi)$ with one backward pass (optimization details in Appendix \ref{sec:implement}).

\section{Sampling and Adaptive Weighting}\label{sec:theory}
We compare four two-layer computations using common parameters, inputs, and first-layer role samples. Dropout is disabled. At the second layer, roles either persist from the first layer or are independently resampled. Local weights use each sample's own endpoint states. Shared weights replace each edge's local weights with their average over samples. All sums below cover incoming edges with distinct endpoints. Appendix \ref{sec:app:feedback} provides the expressions for self-edges and greater depth.

\subsection{Target-state feedback across layers}
For the signed gate $D_{ij}=d_{ij}(Z)$, its mean $\mu_{ij}$ and variance $\nu_{ij}$ are
\begin{equation}
\mu_{ij}=\pi^+_{ij}-\gamma\pi^-_{ij},\qquad
\nu_{ij}=\pi^+_{ij}+\gamma^2\pi^-_{ij}-\mu_{ij}^2.
\label{eq:gate-moments}
\end{equation}
First-layer weights and values agree across samples because they use the common initial state. Role sampling produces sample-specific node representations that determine the second-layer scores.

For an edge $e=(i,j)$ with $i\ne j$, changing its first-layer role changes the target state. The source state is unchanged because all nodes update simultaneously. Let $U_e^s$ be the second-layer weight after setting that role to $s$, and let $V_e=W_v^{(1)}h_j^{(1)}$ be the source value. For a hidden-space direction $w$, define $\eta_e=w^\top W_o^{(1)}V_e$ as the projected source contribution and $d_s^{(\ell)}=\mathbb I(s=+1)-\gamma^{(\ell)}\mathbb I(s=-1)$ as the layer-specific gate. The expected difference between persistent and resampled local-weight preactivations is
\begin{equation}
\mathcal F_i(w)=\sum_{e=(i,j)}\mathbb E_{Z_{-e}}\left[
\eta_e\sum_{s<t}\pi_e^s\pi_e^t
(d_t^{(1)}-d_s^{(1)})(U_e^t-U_e^s)\right].
\label{eq:direct-feedback}
\end{equation}
We order roles as $-1<0<+1$ and write $\pi_e=\pi_{ij}$. We hold the remaining edge roles, collected in $Z_{-e}$, constant during each intervention. Each term in $\mathcal F_i(w)$ combines the gate difference, the corresponding change in the later weight, and the projected source contribution. Their sum measures feedback through the target state. Proposition \ref{prop:target-feedback} states the sign condition and a magnitude bound.

For a source-only scorer, intervening on the edge's own role leaves its weight unchanged, so the feedback is zero. The pairwise expression also applies to finite-valued gates. Binary activity and signed gates each contribute one role pair. The three-state support includes changes in both activity and polarity (Appendix \ref{sec:app:feedback}).

\subsection{Decomposition of the weighting contrast}
Let $U_{e,k}$ be edge $e$'s second-layer weight in sample $k$, with shared weight $\bar U_e=K^{-1}\sum_kU_{e,k}$. Write $\bar a_i^{\tau,\kappa}$ for the sample-average preactivation, where $\tau\in\{\mathrm p,\mathrm r\}$ denotes persistent or resampled roles and $\kappa\in\{\mathrm l,\mathrm s\}$ denotes local or shared weights. We define the weight-source contribution as
\begin{equation}
\mathcal S_i(w)=\sum_{e=(i,j)}\mu_e w^\top W_o^{(1)}\operatorname{Cov}(U_e,V_e).
\end{equation}
Both the random weight $U_e$ and source value $V_e$ depend on the first-layer role draw. Here, $\mu_e=\mathbb E[D_e]$ for the second-layer gate $D_e=d_{z_e}^{(1)}$. Weights and source values can therefore remain correlated when the second-layer gates are independently resampled.

For scalar weights $U_k$ and vector contributions $A_k$, let $\bar U$ and $\bar A$ denote their sample means. Their local and shared averages satisfy
\begin{equation}
\frac1K\sum_kU_kA_k-\bar U\bar A
=\frac1{2K^2}\sum_{k,b}(U_k-U_b)(A_k-A_b).
\label{eq:sample-sharing-identity}
\end{equation}
For independent, identically distributed pairs, the expectation is $(1-1/K)\operatorname{Cov}(U,A)$. Applying this identity to the gate-value products yields the two weighting contrasts:
\begin{equation}
\begin{aligned}
\mathbb E[w^\top(\bar a_i^{\mathrm{r,l}}-\bar a_i^{\mathrm{r,s}})]
&=\left(1-\frac1K\right)\mathcal S_i(w),\\
\mathbb E[w^\top(\bar a_i^{\mathrm{p,l}}-\bar a_i^{\mathrm{p,s}})]
&=\left(1-\frac1K\right)\bigl(\mathcal S_i(w)+\mathcal F_i(w)\bigr).
\end{aligned}
\label{eq:allocation-decomposition}
\end{equation}
Weight-source covariance contributes under both sampling rules; persistence adds target-state feedback. Subtracting the contrasts cancels the covariance contribution and leaves the sampling-weighting interaction:
\begin{equation}
\begin{aligned}
\mathcal J_i(w)
&=\mathbb E\!\left[w^\top\bigl\{(\bar a_i^{\mathrm{p,l}}-\bar a_i^{\mathrm{p,s}})
-(\bar a_i^{\mathrm{r,l}}-\bar a_i^{\mathrm{r,s}})\bigr\}\right]\\
&=\left(1-\frac1K\right)\mathcal F_i(w).
\end{aligned}
\label{eq:graph-interaction}
\end{equation}
The shared mean contains each sample's own weight, so a fraction $1/K$ of its feedback remains:
\begin{equation}
\mathbb E[w^\top(\bar a_i^{\mathrm{p,s}}-\bar a_i^{\mathrm{r,s}})]
=\frac1K\mathcal F_i(w).
\label{eq:shared-feedback}
\end{equation}
Local and shared weights coincide at $K=1$. For $K>1$, their preactivation ordering reverses between sampling rules exactly when $\mathcal S_i(w)[\mathcal S_i(w)+\mathcal F_i(w)]<0$. The feedback must oppose and exceed the weight-source contribution in magnitude. A source-only scorer can produce a weighting contrast through $\mathcal S_i(w)$, and its persistence interaction is zero. The finite-margin identity in Appendix \ref{sec:app:margin} describes the corresponding class-probability changes after activation and classification.

\subsection{Distribution-preserving weight reassignment}
For $K\ge2$, let $\sigma$ be a uniformly sampled permutation of $\{1,\ldots,K\}$ with $\sigma(k)\ne k$ for every $k$, independent of the role draws. We assign $U_{e,\sigma(k)}$ to sample $k$ on every edge and keep that sample's gates, source values, and self-contribution. Applying the same permutation to complete weight vectors preserves their empirical distribution and changes their pairing with the sampled states.

Let $\bar a_i^{\tau,\sigma}$ be the sample-average preactivation after reassignment. Averaging over permutations yields
\begin{equation}
\bar a_i^{\tau,\mathrm l}-\mathbb E_\sigma[\bar a_i^{\tau,\sigma}]
=\frac K{K-1}\bigl(\bar a_i^{\tau,\mathrm l}-\bar a_i^{\tau,\mathrm s}\bigr).
\label{eq:permutation-batch}
\end{equation}
Averaging over role batches recovers the full covariance contrast, so the permutation-based persistence interaction equals $\mathcal F_i(w)$. Equation \ref{eq:permutation-batch} relates the contrast from weight sharing to the contrast from reassigning the original vectors (Appendix \ref{sec:app:permutation}).

\subsection{Analytical verification and depth dependence}
In Appendix \ref{sec:app:feature_coupling_design}, we vary sample count and scorer response in a three-node ReLU construction. Exact enumeration reproduces the interaction's sign and zero cases, as well as the changes in predictive probability and mixture loss (Tables \ref{tab:relu-factorial} and \ref{tab:relu-factorial-response}). We also verify the decomposition and source-only identity on a vector-valued graph (Table \ref{tab:theory-verification}). These finite-state checks isolate the fixed-parameter mechanism; the ordering after independent retraining is evaluated separately in Section \ref{sec:experiments}.

At greater depth, let $d_e$ be the shortest directed distance from target $i$ to source $j$, or infinity when a return path is absent. The source value entering layer $L$ is independent of the edge's role when $L<d_e+2$ (Proposition \ref{prop:return-path}). The earlier role can reach the source state through a directed path from the target. Once $L\ge d_e+2$, the role may also affect the source state, introducing an additional source-response term.

\begin{table*}[t]
\caption{Feature-only transductive node classification using node features and graph structure on shared partitions. Entries are means and sample standard deviations over ten runs. Minesweeper uses ROC-AUC, and the other graphs use accuracy. Boldface marks the highest mean. The accompanying code specifies the evaluated variants marked by $\dagger$.}
\label{tab:hetero_bench}
\centering
\setlength{\tabcolsep}{3pt}
\begin{adjustbox}{width=\textwidth}
\begin{tabular}{@{}l|cccccccc@{}}
\Xhline{2\arrayrulewidth}
\textbf{Dataset}
& \textbf{Roman.}
& \textbf{Mine.}
& \textbf{Amazon.}
& \textbf{Chameleon}
& \textbf{Squirrel}
& \textbf{Actor}
& \textbf{Cornell}
& \textbf{Wisconsin} \\
$\mathcal{G}_h$ (Eq. \ref{global_homophily})
& 0.05 & 0.68 & 0.38 & 0.23 & 0.22 & 0.22 & 0.30 & 0.21 \\
\Xhline{2\arrayrulewidth}

GCN
& 57.3$_{\pm 0.65}$ & 81.4$_{\pm 0.98}$ & 38.5$_{\pm 0.45}$
& 54.9$_{\pm 0.59}$ & 31.1$_{\pm 0.71}$ & 24.5$_{\pm 0.58}$
& 49.6$_{\pm 3.28}$ & 49.0$_{\pm 2.59}$ \\

GAT
& 56.6$_{\pm 0.51}$ & 80.0$_{\pm 1.08}$ & 39.0$_{\pm 0.52}$
& 54.4$_{\pm 0.84}$ & 31.0$_{\pm 0.93}$ & 28.2$_{\pm 0.46}$
& 52.4$_{\pm 2.62}$ & 50.2$_{\pm 3.05}$ \\

H\textsubscript{2}GCN
& 60.6$_{\pm 0.54}$ & 84.9$_{\pm 1.30}$ & 41.3$_{\pm 0.62}$
& 53.1$_{\pm 0.88}$ & 31.2$_{\pm 0.68}$ & 25.9$_{\pm 1.07}$
& 54.9$_{\pm 2.87}$ & 62.0$_{\pm 3.04}$ \\

GCNII
& 69.2$_{\pm 0.48}$ & 84.8$_{\pm 1.35}$ & 41.6$_{\pm 0.59}$
& 54.0$_{\pm 0.77}$ & 30.8$_{\pm 0.91}$ & 26.2$_{\pm 2.22}$
& 55.9$_{\pm 3.03}$ & 63.9$_{\pm 3.46}$ \\

MagNet
& 65.2$_{\pm 0.64}$ & 85.6$_{\pm 1.48}$ & 41.9$_{\pm 0.71}$
& 56.9$_{\pm 1.34}$ & 32.4$_{\pm 1.15}$ & 26.4$_{\pm 0.97}$
& 55.1$_{\pm 3.40}$ & 61.7$_{\pm 3.08}$ \\

GPRGNN
& 63.1$_{\pm 0.60}$ & 85.3$_{\pm 1.19}$ & 42.0$_{\pm 0.63}$
& 55.8$_{\pm 0.81}$ & 30.6$_{\pm 0.63}$ & 25.2$_{\pm 0.89}$
& 51.4$_{\pm 2.80}$ & 63.1$_{\pm 2.60}$ \\

FAGCN
& 61.7$_{\pm 0.66}$ & 83.5$_{\pm 1.26}$ & 40.9$_{\pm 0.59}$
& 54.8$_{\pm 0.81}$ & 31.2$_{\pm 0.87}$ & 26.8$_{\pm 1.24}$
& 56.8$_{\pm 3.87}$ & 64.3$_{\pm 3.19}$ \\

ACM-GCN
& 64.5$_{\pm 0.67}$ & 86.1$_{\pm 1.34}$ & 42.4$_{\pm 0.61}$
& 56.6$_{\pm 1.40}$ & 32.1$_{\pm 1.05}$ & 25.9$_{\pm 2.02}$
& 55.1$_{\pm 3.35}$ & 62.1$_{\pm 3.65}$ \\

GloGNN
& 70.2$_{\pm 0.64}$ & 85.7$_{\pm 1.27}$ & 41.9$_{\pm 0.67}$
& 53.9$_{\pm 0.70}$ & 31.0$_{\pm 0.82}$ & 27.0$_{\pm 0.73}$
& 48.6$_{\pm 4.25}$ & 60.2$_{\pm 4.12}$ \\

Auto-HeG$^{\dagger}$
& 66.3$_{\pm 0.67}$ & 86.0$_{\pm 1.38}$ & 42.7$_{\pm 0.68}$
& 54.3$_{\pm 1.33}$ & 31.7$_{\pm 1.11}$ & 26.5$_{\pm 0.99}$
& 53.8$_{\pm 4.02}$ & 64.0$_{\pm 4.49}$ \\

GAT-sep
& 72.5$_{\pm 0.42}$ & 85.6$_{\pm 0.63}$ & 44.5$_{\pm 0.50}$
& 59.0$_{\pm 1.84}$ & 34.2$_{\pm 0.70}$ & 34.0$_{\pm 0.92}$
& 55.1$_{\pm 2.66}$ & 70.2$_{\pm 3.84}$ \\

DirGNN
& 67.1$_{\pm 0.69}$ & 86.2$_{\pm 1.46}$ & 43.4$_{\pm 0.71}$
& 59.8$_{\pm 1.45}$ & 35.2$_{\pm 1.13}$ & 27.5$_{\pm 0.95}$
& 63.8$_{\pm 2.80}$ & 63.0$_{\pm 4.23}$ \\

PCNet
& 64.2$_{\pm 0.64}$ & 85.9$_{\pm 1.32}$ & 42.3$_{\pm 0.64}$
& 57.6$_{\pm 1.65}$ & 31.8$_{\pm 0.58}$ & 26.6$_{\pm 0.90}$
& 54.1$_{\pm 3.02}$ & 60.4$_{\pm 3.42}$ \\

TFE-GNN$^{\dagger}$
& 68.7$_{\pm 0.70}$ & 86.1$_{\pm 1.50}$ & 43.7$_{\pm 0.72}$
& 60.2$_{\pm 1.61}$ & \textbf{36.0$_{\pm 0.59}$} & 28.1$_{\pm 0.81}$
& 53.7$_{\pm 4.07}$ & 62.5$_{\pm 3.99}$ \\

CGNN
& 70.3$_{\pm 0.75}$ & 86.6$_{\pm 1.53}$ & 43.9$_{\pm 0.75}$
& 59.1$_{\pm 0.78}$ & 34.4$_{\pm 0.97}$ & 32.4$_{\pm 1.17}$
& 62.4$_{\pm 3.25}$ & 64.9$_{\pm 5.22}$ \\

L2DGCN
& 65.4$_{\pm 0.65}$ & 85.7$_{\pm 1.37}$ & 43.2$_{\pm 0.67}$
& 53.1$_{\pm 0.37}$ & 35.4$_{\pm 0.52}$ & 35.3$_{\pm 0.35}$
& 61.6$_{\pm 3.28}$ & 65.8$_{\pm 5.01}$ \\

\hline
\rowcolor{yellow!20}
\textbf{P3MP (feat.)}
& \textbf{73.6$_{\pm 0.98}$} & \textbf{86.8$_{\pm 1.03}$} & \textbf{45.3$_{\pm 0.71}$} & \textbf{62.1$_{\pm 1.12}$} & 35.6$_{\pm 0.45}$ & \textbf{38.0$_{\pm 0.77}$} & \textbf{69.0$_{\pm 2.83}$} & \textbf{71.0$_{\pm 2.43}$} \\

\Xhline{2\arrayrulewidth}
\end{tabular}
\end{adjustbox}
\end{table*}

\section{Experiments}\label{sec:experiments}
\subsection{Feature-only node classification}
We train all methods on the same ten partitions per graph, with one initialization per partition. Training, validation, and test fractions are 10\%, 25\%, and 65\%, or 20\%, 40\%, and 40\% on Cornell and Wisconsin. Checkpoint selection and test evaluation use ROC-AUC on Minesweeper and accuracy on the other graphs. We report means and sample standard deviations over runs. P3MP uses a feature MLP on legacy Chameleon and Squirrel, which include their documented duplicate nodes \citep{platonov2023critical}, and a two-layer single-head GAT elsewhere. Appendix \ref{sec:implement} provides the architecture and evaluation details. These scores are specific to the stated splits and preprocessing. P3MP has the highest mean score on seven of the eight graphs in Table \ref{tab:hetero_bench}; TFE-GNN leads on Squirrel by 0.4 percentage points. P3MP also has higher mean accuracy than selective blocking \citep{choi2025selective} and GCN-BBGDC \citep{hasanzadeh2020bayesian} in Table \ref{tab:matched_routing_thresholding}. Because selective blocking and GCN-BBGDC use different architectures, interpret these results as method-level comparisons, not controlled component-wise ablations.

\subsection{Persistence and weight sharing}
We train the four feature-only sampling-weighting combinations independently. Each uses mixture NLL with the same backbone and scorer architectures, partitions, sample counts, and optimization settings. Shared weights enter both message propagation and the active-weight penalty.

For dataset $d$ and split $t$, let $A_{d,t}^{\tau,\kappa}$ and $N_{d,t}^{\tau,\kappa}$ denote accuracy and NLL, using the indices from Section \ref{sec:theory}. The difference in the benefit of local weights is
\begin{equation}
\begin{aligned}
\Delta_{d,t}^{\mathrm{Acc}}&=(A_{d,t}^{\mathrm{p,l}}-A_{d,t}^{\mathrm{p,s}})
-(A_{d,t}^{\mathrm{r,l}}-A_{d,t}^{\mathrm{r,s}}),\\
\Delta_{d,t}^{\mathrm{NLL}}&=(N_{d,t}^{\mathrm{p,s}}-N_{d,t}^{\mathrm{p,l}})
-(N_{d,t}^{\mathrm{r,s}}-N_{d,t}^{\mathrm{r,l}}).
\end{aligned}
\label{eq:feature-factorial-interaction}
\end{equation}
Positive values indicate a larger benefit from local weighting under persistence. We compute each contrast within a split and then summarize over splits (Appendix \ref{sec:app:controls}). In Table \ref{tab:feature_factorial}, local weights yield higher accuracy and lower NLL under persistence, and shared weights yield higher accuracy and lower NLL under resampling. This reversal in the preferred weighting rule also appears in accuracy on deduplicated Chameleon and Squirrel (Appendix \ref{app:deduplicated_persistence}).

\begin{table}[ht]
\centering
\caption{Feature-only persistence-by-weighting results over ten shared splits. Entries report mean $\pm$ sample standard deviation. Interaction rows summarize the splitwise paired contrasts in Eq. \ref{eq:feature-factorial-interaction}, with accuracy interactions in percentage points.}
\label{tab:feature_factorial}
\small
\setlength{\tabcolsep}{3pt}
\renewcommand{\arraystretch}{1.08}
\begin{tabular*}{\linewidth}{@{\extracolsep{\fill}}llccccc@{}}
\toprule
Sampling & Weights & Roman. & Cham. & Squirrel & Actor & Cornell \\
\midrule
\rowcolor{yellow!20}
\multicolumn{7}{@{}l}{\textbf{Test accuracy (\%, mean $\pm$ std)}} \\
Persistent & Local & $73.6_{\pm 0.98}$ & $62.1_{\pm 1.12}$ & $35.6_{\pm 0.45}$ & $38.0_{\pm 0.77}$ & $69.0_{\pm 2.83}$ \\
Persistent & Shared & $72.8_{\pm 1.20}$ & $61.0_{\pm 1.47}$ & $35.1_{\pm 0.84}$ & $36.0_{\pm 1.10}$ & $68.4_{\pm 3.49}$ \\
Resampled & Local & $71.4_{\pm 0.91}$ & $60.5_{\pm 1.05}$ & $34.0_{\pm 0.58}$ & $35.2_{\pm 0.82}$ & $67.6_{\pm 3.95}$ \\
Resampled & Shared & $71.8_{\pm 1.01}$ & $61.1_{\pm 1.20}$ & $34.6_{\pm 0.42}$ & $35.7_{\pm 1.00}$ & $68.4_{\pm 3.17}$ \\
\multicolumn{2}{@{}l}{Interaction (percentage points)} & $1.2_{\pm 0.6}$ & $1.7_{\pm 0.9}$ & $1.1_{\pm 0.3}$ & $2.5_{\pm 0.6}$ & $1.4_{\pm 0.6}$ \\
\midrule
\rowcolor{yellow!20}
\multicolumn{7}{@{}l}{\textbf{NLL (mean $\pm$ std)}} \\
Persistent & Local & $1.067_{\pm 0.062}$ & $0.976_{\pm 0.048}$ & $1.420_{\pm 0.026}$ & $1.400_{\pm 0.029}$ & $0.826_{\pm 0.073}$ \\
Persistent & Shared & $1.084_{\pm 0.067}$ & $0.996_{\pm 0.051}$ & $1.431_{\pm 0.027}$ & $1.410_{\pm 0.030}$ & $0.851_{\pm 0.084}$ \\
Resampled & Local & $1.142_{\pm 0.052}$ & $1.007_{\pm 0.047}$ & $1.442_{\pm 0.023}$ & $1.425_{\pm 0.027}$ & $0.865_{\pm 0.059}$ \\
Resampled & Shared & $1.132_{\pm 0.084}$ & $0.991_{\pm 0.066}$ & $1.434_{\pm 0.058}$ & $1.419_{\pm 0.014}$ & $0.848_{\pm 0.103}$ \\
\multicolumn{2}{@{}l}{NLL interaction} & $0.027_{\pm 0.01}$ & $0.036_{\pm 0.03}$ & $0.019_{\pm 0.01}$ & $0.016_{\pm 0.01}$ & $0.042_{\pm 0.02}$ \\
\bottomrule
\end{tabular*}
\end{table}

\begin{table}[t]
\centering
\caption{Feature-only sampling comparisons over ten shared splits. Entries are accuracy means and standard deviations (\%). Internal controls share P3MP's backbone and scorer. Selective blocking and GCN-BBGDC use their own architectures. Boldface marks the highest mean.}
\label{tab:matched_routing_thresholding}
\small
\setlength{\tabcolsep}{3pt}
\renewcommand{\arraystretch}{1.12}
\begin{tabular*}{\linewidth}{@{\extracolsep{\fill}}lccccc@{}}
\toprule
Configuration & Roman. & Cham. & Squirrel & Actor & Cornell \\
\midrule
\rowcolor{yellow!20}
\textbf{Full P3MP (feat.)}
& \textbf{73.6}$_{\pm 0.98}$
& \textbf{62.1}$_{\pm 1.12}$
& \textbf{35.6}$_{\pm 0.45}$
& \textbf{38.0}$_{\pm 0.77}$
& \textbf{69.0}$_{\pm 2.83}$ \\
Layerwise resampling
& 71.4$_{\pm 0.91}$ & 60.5$_{\pm 1.05}$ & 34.0$_{\pm 0.58}$ & 35.2$_{\pm 0.82}$ & 67.6$_{\pm 3.95}$ \\
Soft gating, retrained
& 72.2$_{\pm 1.02}$ & 61.0$_{\pm 1.18}$ & 34.5$_{\pm 0.42}$ & 35.5$_{\pm 0.71}$ & 68.0$_{\pm 2.12}$ \\
Threshold zero, same scorer
& 73.3$_{\pm 0.85}$ & 60.8$_{\pm 1.09}$ & 34.7$_{\pm 0.66}$ & 35.8$_{\pm 0.74}$ & 68.1$_{\pm 1.88}$ \\
\midrule
Selective blocking
& 65.4$_{\pm 0.81}$ & 54.1$_{\pm 0.52}$ & 31.6$_{\pm 0.51}$ & 28.0$_{\pm 0.72}$ & 58.1$_{\pm 3.91}$ \\
GCN-BBGDC
& 63.8$_{\pm 2.06}$ & 54.0$_{\pm 1.95}$ & 30.8$_{\pm 0.82}$ & 27.4$_{\pm 1.02}$ & 48.6$_{\pm 5.23}$ \\
\bottomrule
\end{tabular*}
\end{table}

Persistent sampling also improves accuracy over layerwise resampling and retrained soft gating in Table \ref{tab:matched_routing_thresholding}. Positive thresholding improves accuracy over the zero-threshold control, which uses the same scorer and penalty weight with $u_{ij}=|s_{ij}|$. Thresholding reduces each weight by $\min\{|s_{ij}|,\lambda_{\mathrm{sc}}\}$, changing both the weight magnitude and message support.

\paragraph{Responses at learned checkpoints.}
At feature-only P3MP checkpoints, we compare the coupled second-layer computations with the learned parameters and first-layer states unchanged. We project messages along the target-class classifier direction relative to the mean competing class. In Table \ref{tab:frozen_intervention}, the mean interaction is positive for the endpoint scorer and close to zero when its target projection is set to zero, as predicted for a source-only scorer. For each checkpoint, we average paired measurements over 100 independent batches of five role samples. Test labels specify the projection direction; the structural encoder receives features alone. Appendix \ref{sec:app:frozen_intervention} describes the procedure.

\begin{table}[ht]
\centering
\caption{Feature-only checkpoint interactions, reported as means $\pm$ sample standard deviations across ten splits after averaging the paired contrasts within each checkpoint.}
\label{tab:frozen_intervention}
\small
\setlength{\tabcolsep}{6pt}
\renewcommand{\arraystretch}{1.05}
\begin{tabular*}{\linewidth}{@{\extracolsep{\fill}}lcc@{}}
\toprule
Dataset & Endpoint scorer & Source-only scorer \\
\midrule
Roman-Empire & $0.13_{\pm 0.03}$ & $0.002_{\pm 0.008}$ \\
Chameleon    & $0.07_{\pm 0.02}$ & $-0.002_{\pm 0.005}$ \\
Squirrel     & $0.05_{\pm 0.02}$ & $0.003_{\pm 0.006}$ \\
Actor        & $0.12_{\pm 0.04}$ & $0.001_{\pm 0.010}$ \\
Cornell      & $0.09_{\pm 0.05}$ & $-0.004_{\pm 0.013}$ \\
\bottomrule
\end{tabular*}
\end{table}

\subsection{Label-aware extension}
\begin{figure}[!ht]
\centering
\begin{subfigure}[t]{0.485\textwidth}
\centering
\includegraphics[width=\linewidth]{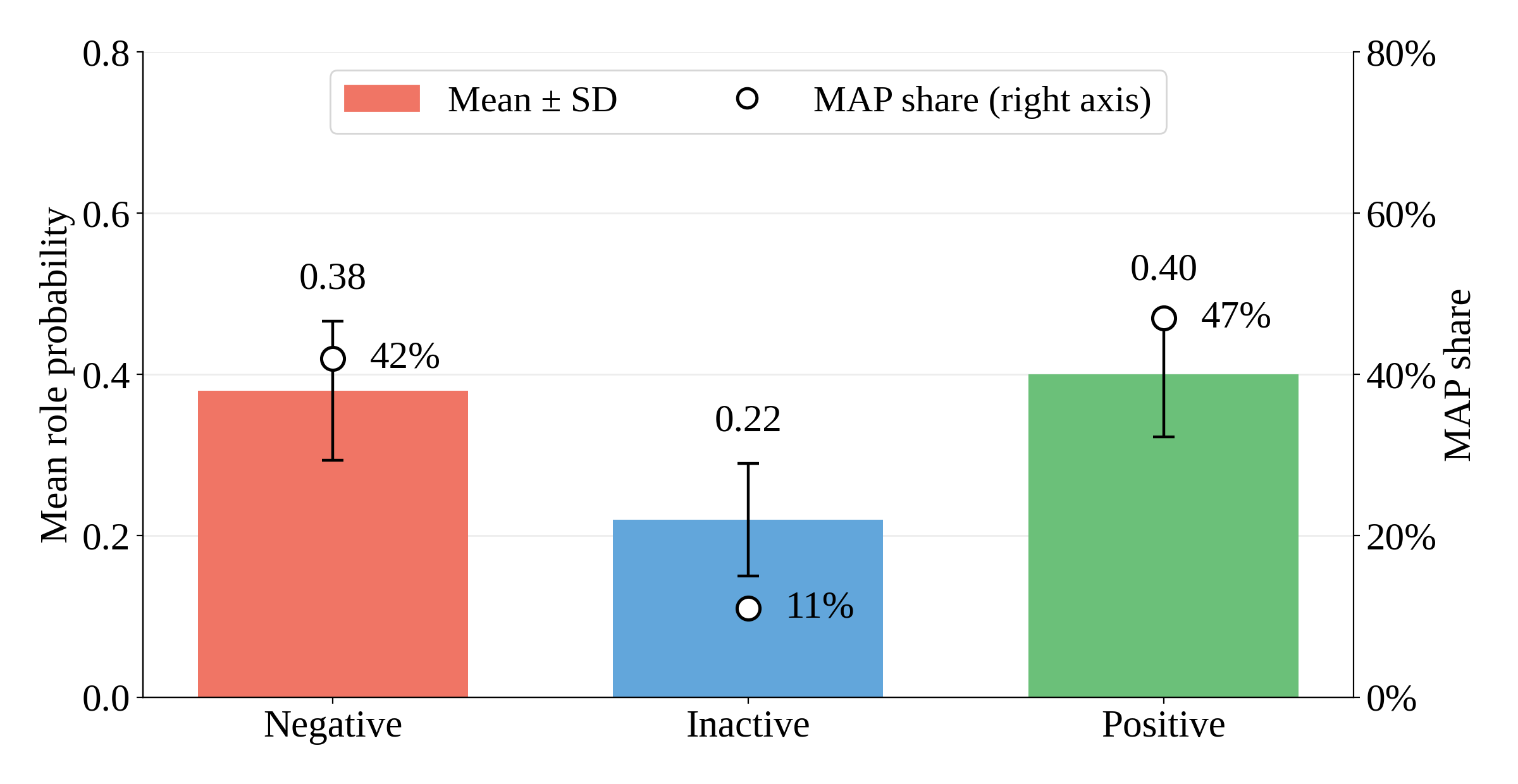}
\caption{Edge-role statistics on Chameleon}
\label{fig:post}
\end{subfigure}
\hfill
\begin{subfigure}[t]{0.485\textwidth}
\centering
\includegraphics[width=\linewidth]{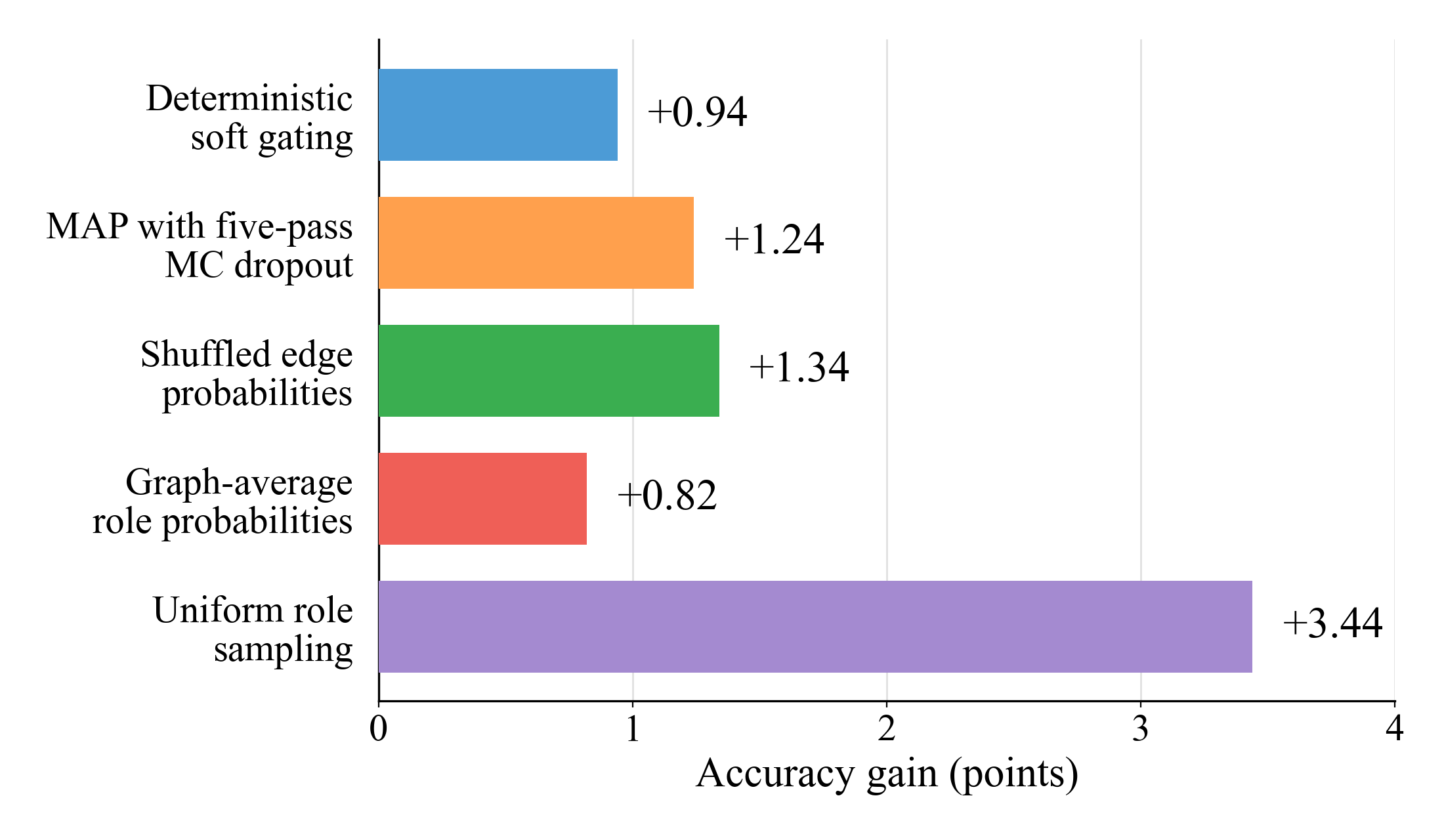}
\caption{Prediction-time accuracy contrasts}
\label{fig:fixed_controls}
\end{subfigure}
\smallskip
\begin{subfigure}[t]{0.485\textwidth}
\centering
\includegraphics[width=\linewidth]{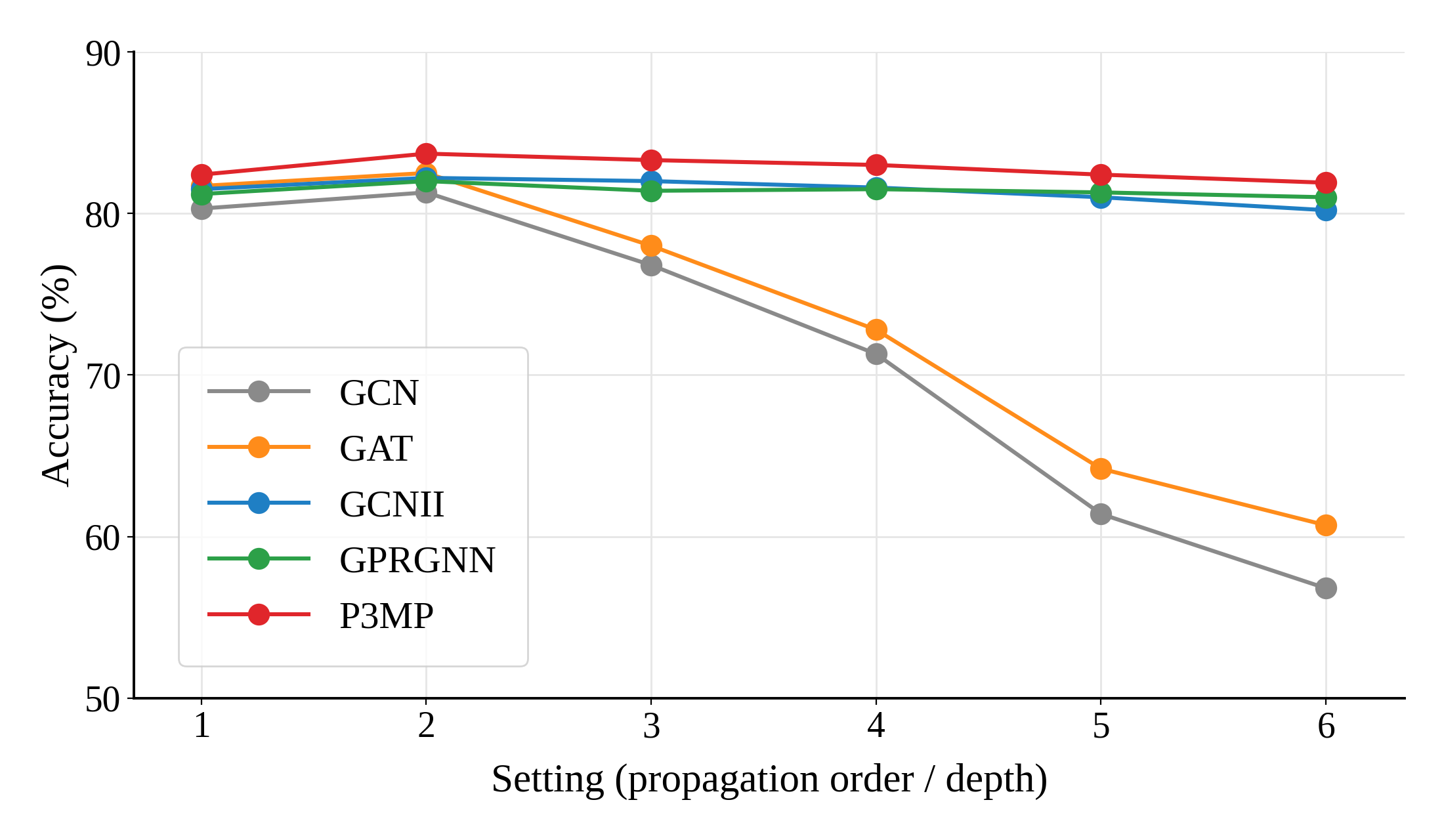}
\caption{Signed depth and propagation order on Cora}
\label{fig:depth}
\end{subfigure}
\hfill
\begin{subfigure}[t]{0.485\textwidth}
\centering
\includegraphics[width=\linewidth]{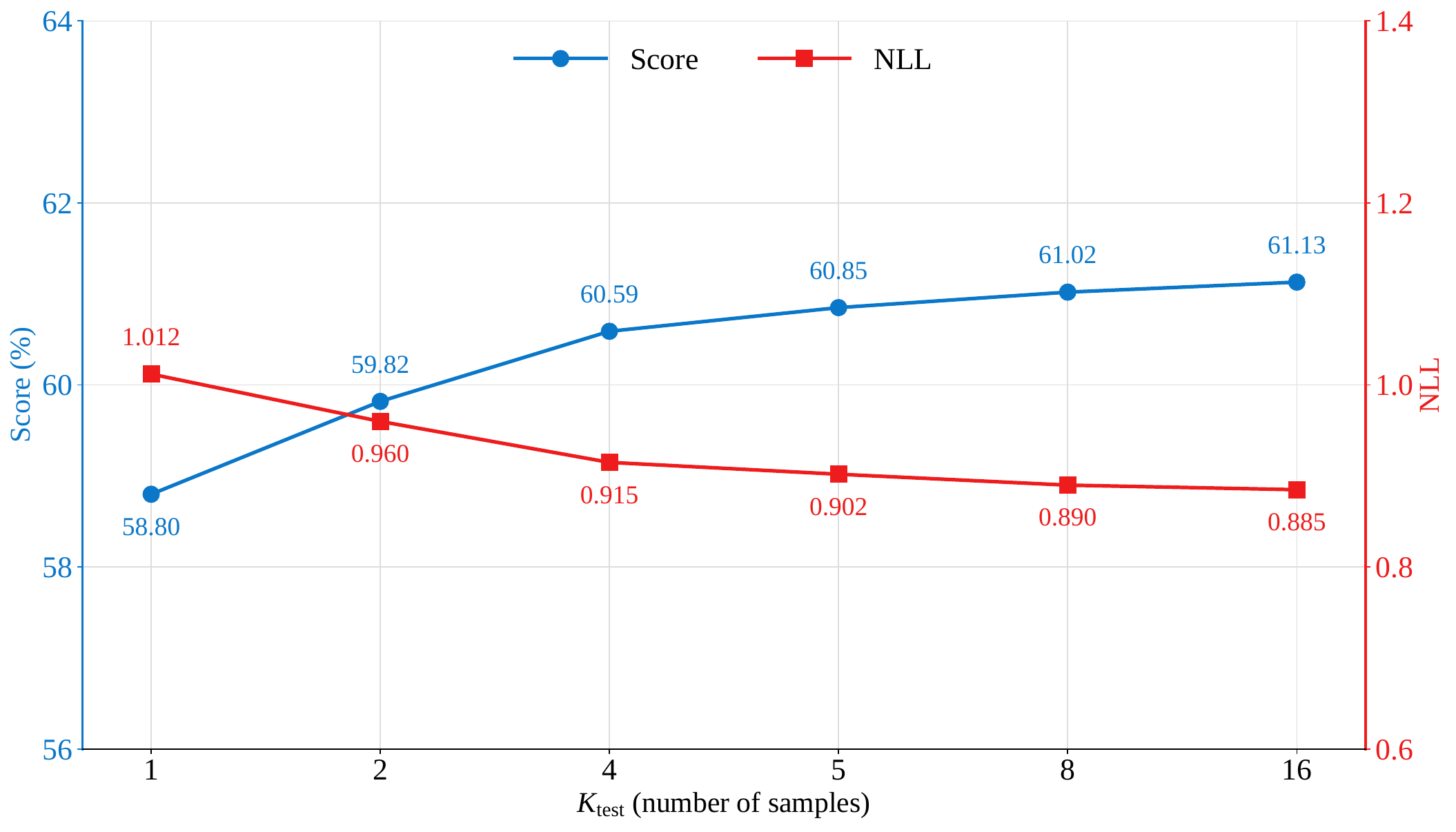}
\caption{Prediction sample count}
\label{fig:sample_cost}
\end{subfigure}
\caption{Diagnostics for the label-aware extension. (\textbf{a}) Chameleon role probabilities and MAP shares. (\textbf{b}) Accuracy contrasts across five graphs with five checkpoints each. (\textbf{c}) Cora accuracy against P3MP signed depth after its backbone, comparison-model layer count, or GPRGNN propagation order. (\textbf{d}) Primary score on the left axis and NLL on the right axis, averaged across eight graphs with five checkpoints each. }
\label{fig:diagnostics}
\end{figure}

\begin{figure}[ht]
    \centering
    \includegraphics[width=\linewidth]{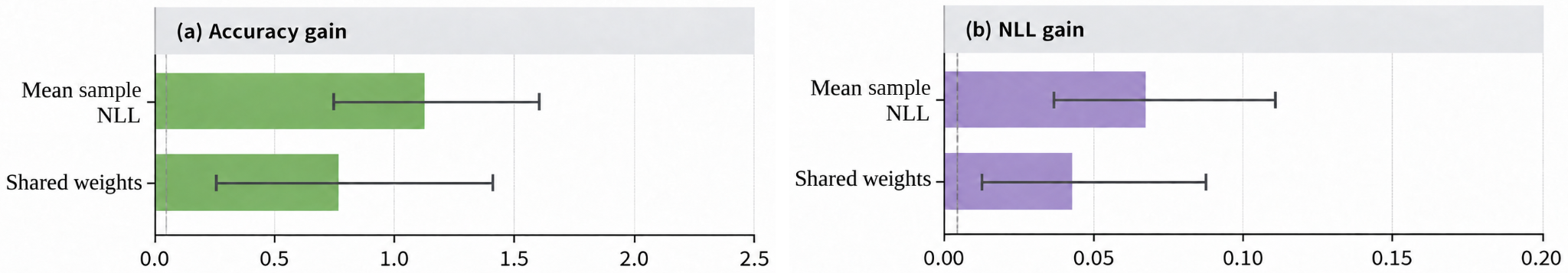}
    \caption{Label-aware retraining contrasts for mean sample NLL and shared weights. Accuracy (a) is full model minus control, and NLL (b) is control minus full model, so positive values indicate gains for P3MP. Bars show means over five datasets and ten paired runs. Error bars give 95\% intervals.}
    \label{fig:matched_effects}
\end{figure}

The extension adds training-label indicators to the structural encoder's features. Each supervised node supplies its own indicator, and validation and test rows are zero. Training-label indicators condition the role distribution but are not provided as inputs to the prediction backbone. The diagnostics below concern this label-aware setting, whereas the main sampling-weighting comparisons and checkpoint interventions use feature-only inputs. Figures \ref{fig:diagnostics} and \ref{fig:matched_effects} report the role distributions and predictive-averaging comparisons for this model. Appendix \ref{sec:label_extension} defines the input, and Table \ref{tab:evaluation_groups} lists the graph sets and run counts.

Figure \ref{fig:post} reports the means, standard deviations, and maximum a posteriori (MAP) shares of Chameleon's role probabilities. The additive and subtractive means are similar, and both exceed the inactive mean, as favored by the structural regularizer.

Probability averaging gives higher accuracy than soft gating or MAP with Monte Carlo dropout in Figure \ref{fig:fixed_controls}. Learned edge-specific probabilities also improve accuracy over shuffling, graph-average replacement, and uniform sampling. Graph-average replacement assigns the mean role vector $\bar\pi=m^{-1}\sum_{(i,j)\in\mathcal E_{\mathrm{obs}}}\pi_{ij}$ to every edge.

In Figure \ref{fig:depth}, we vary P3MP's signed depth after an unchanged backbone. For comparison, we vary the layer count of GCN, GAT, and GCNII or the propagation order of GPRGNN. Accuracy is more stable for P3MP, GCNII, and GPRGNN as depth or order increases.

Mixture NLL and local weights yield higher accuracy and lower NLL than their respective retrained controls in Figure \ref{fig:matched_effects}. Every model averages five predictions. Let $\boldsymbol u_K=(1/K,\ldots,1/K)$ be the uniform sample distribution and $\boldsymbol r_i=(r_{i1},\ldots,r_{iK})$ the responsibilities from Eq. \ref{eq:responsibilities}. The objective gap is
\begin{equation}
\mathcal L_{\mathrm{sample},i}-\mathcal L_{\mathrm{cls},i}
=\mathrm{KL}(\boldsymbol u_K\|\boldsymbol r_i),\qquad
\mathcal L_{\mathrm{sample},i}=-\frac1K\sum_k\log P_{ik}.
\label{eq:objective-gap}
\end{equation}
The additional KL term favors uniform responsibilities. Mixture NLL trains the predictive average using sample contributions that can differ between target nodes (Appendix \ref{sec:app:mixture}). In the role-support comparisons, tri-state roles yield higher accuracy than signed-only and activity-only roles. Accuracy also improves with the active-weight penalty and both structural regularizers (Table \ref{tab:component_ablation}).

\paragraph{Prediction sample count.}
Figure \ref{fig:sample_cost} reports the primary score and NLL averaged over eight graphs, using checkpoints trained with five samples. The score combines seven accuracies and Minesweeper ROC-AUC. The score increases, and NLL decreases with the prediction sample count. The score is 60.85 at the training count $K=5$ and reaches 61.13 at $K=16$. Let $\bar p_i=\mathbb E[p_i(Z)]$ be the population predictive vector and $\ell_{i,K}=\mathbb E[-\log\hat p_{i,y_i}^{(K)}]$ the expected mixture NLL. Standard Monte Carlo and sample-average log bounds \citep{burda2015importance} yield
\begin{equation}
\mathbb E\|\hat p_i^{(K)}-\bar p_i\|_2^2
=\frac1K\sum_c\operatorname{Var}(p_{i,c}(Z)),\qquad
-\log\bar p_{i,y_i}\le\ell_{i,K+1}\le\ell_{i,K}.
\end{equation}
Appendix \ref{sec:app:mc} gives the derivations and a class-decision stability bound.

We evaluate the encoders once and run the signed computations sequentially over samples. On Chameleon, latency rises with sample count, reaching 8.17 times the single-sample value at 16 samples; peak memory rises by 0.13\% (Table \ref{tab:sample_resources}). Sequential execution holds one sample's signed states at a time. Stored class log-probabilities contribute an $O(KnC)$ storage term. Appendix \ref{sec:app:cost} provides the measurements, and Eq. \ref{eq:complexity} states the operation count.

\section{Conclusion}
Persistent sampling couples an edge's role to the adaptive weight computed from the node state changed by its earlier message. In the coupled two-layer analysis, target-state feedback and weight-source covariance determine the local versus shared preactivation ordering under each sampling rule. Shared weights retain a $1/K$ fraction of the feedback. Permuting complete weight vectors preserves their empirical distribution while changing their pairing with sampled states. In independently trained feature-only models, local weights yield higher accuracy under persistence, and shared weights yield higher accuracy under resampling. The same ordering holds on deduplicated Chameleon and Squirrel.

\clearpage
\section*{AI Use Statement}
We used generative AI tools to assist with language editing and to help verify selected mathematical arguments. The authors developed the key proof ideas. The authors independently reviewed and verified all mathematical claims and proof steps and take full responsibility for the final manuscript.

\subsection*{Ethics statement}
We evaluate transductive node classification on the benchmarks in Appendix \ref{sec:app:base}. Training labels provide classification supervision, and the label-aware extension uses structural indicators. Validation labels guide model selection, and test labels enter final evaluation and post-hoc annotation diagnostics. Additive and subtractive roles describe computational contributions to class predictions. The study concerns these predictive quantities, with privacy, group-specific errors, and social or causal interpretations outside its evaluation scope. 

\subsection*{Reproducibility statement}
Section \ref{sec:method} defines the model and training objective. Appendix \ref{sec:implement} describes P3MP's architecture, training algorithm, preprocessing, partitions, optimization, and random-number settings, while Appendix \ref{sec:app:controls} defines the experimental controls and paired summaries. The analytical appendix provides proofs and finite-state evaluation formulas.

\bibliographystyle{plainnat}
\bibliography{iclr2027_conference}

\clearpage
\appendix

\section{Illustrative Example}
\begin{figure}[!ht]
\centering
\includegraphics[width=\linewidth]{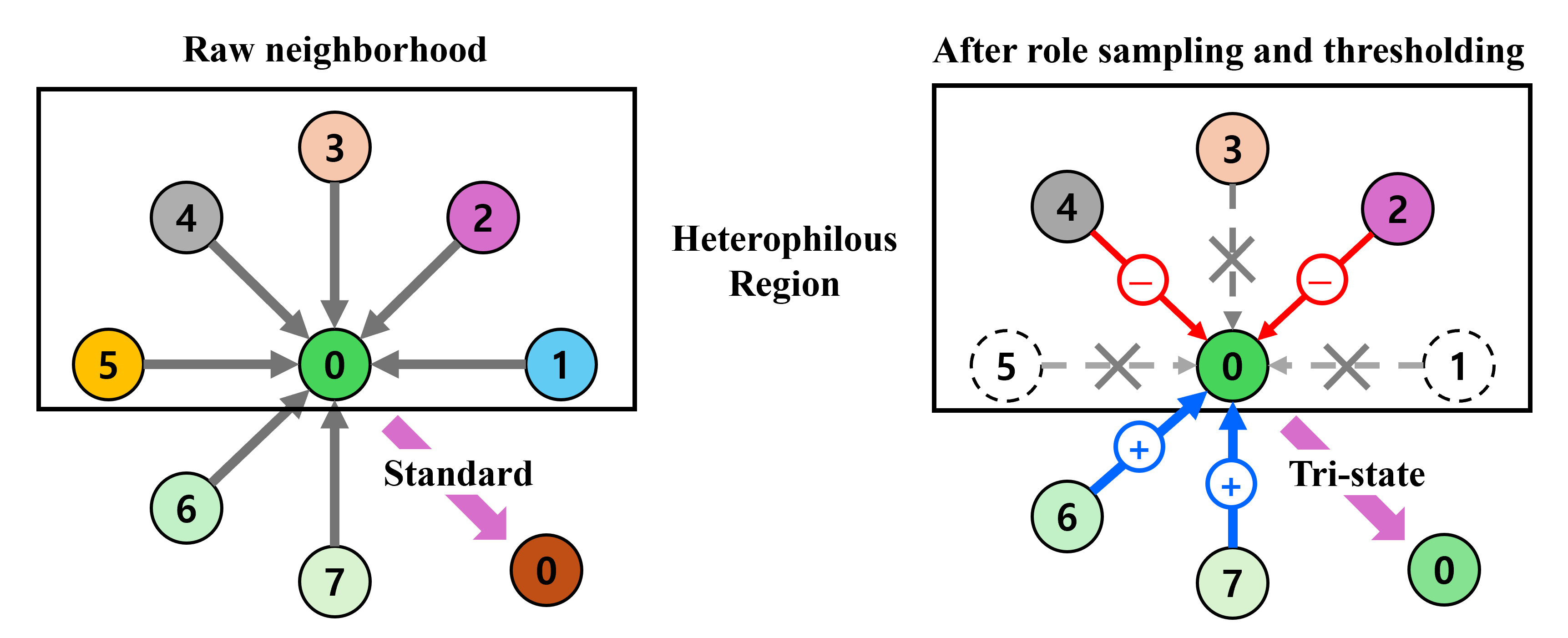}
\caption{Illustrative tri-state assignment around target node 0. The left panel shows standard aggregation over the observed incoming neighborhood. The right panel shows one sampled tri-state update, with additive roles on edges from nodes 6 and 7, subtractive roles on edges from nodes 2 and 4, and inactive roles on edges from nodes 1, 3, and 5. Magenta arrows indicate the schematic aggregated and signed-update states.}
\label{fig:intro_example}
\end{figure}

Figure \ref{fig:intro_example} shows one tri-state draw for the seven incoming edges of target node 0. The left panel depicts the observed neighborhood, and the right panel shows the sampled message operations. The box encloses neighbors whose colors differ from the target's green state. Active edges contribute to the signed aggregate below. Crossed edges have zero contribution for this sample.

Suppressing layer and sample indices, we write Eq. \ref{eq:signed-aggregation} for the right panel as
\begin{equation}
h_0'=W_o\bigl(u_{06}v_6+u_{07}v_7-\gamma u_{02}v_2-\gamma u_{04}v_4\bigr)
+W_{\mathrm{self}}h_0+b.
\end{equation}
The weights $u_{0j}$ and source values $v_j$ use the current endpoint states, and $\gamma$ is the layer's learned subtractive strength. The first four terms apply the sampled signs and thresholded magnitudes to active-edge messages. The final terms add the target's learned self-contribution. The updated states then compute the next layer's weights.

\section{Related Work}
\paragraph{Signed messages and edge selection.}
Graph convolution and attention learn neighborhood transformations and aggregation weights \citep{defferrard2016convolutional,kipf2017semi,velivckovic2018graph}. Signed graph convolution and attention introduce positive and negative relational channels \citep{derr2018signed,huang2019signed}. For multiclass graphs, signed-message analysis relates these contributions to class margins \citep{choi2025beyond}. Selective blocking suppresses messages using local homophily and classification uncertainty \citep{choi2025selective}. P3MP assigns a distribution to additive, subtractive, and inactive operations and computes their nonnegative magnitudes from current endpoint states.

\paragraph{Stochastic connectivity and predictive averaging.}
DropEdge distinguishes adjacency perturbations shared across layers from independent layerwise perturbations \citep{rong2019dropedge}. Bayesian graph convolutional networks average over sampled graph structures \citep{zhang2019bayesian}, and adaptive connection sampling learns connectivity probabilities jointly with a predictor \citep{hasanzadeh2020bayesian}. We analyze these sampling choices when we compute weights from the sampled representations. The decomposition separates target-state feedback from weight-source covariance under cross-sample averaging. Mixture training uses the predictive-average formulation, and sample-average log bounds describe its expected loss as the prediction sample count grows \citep{burda2015importance}.

\paragraph{Neighborhood organization and graph filtering.}
H\textsubscript{2}GCN separates self and neighbor information, Geom-GCN organizes geometric neighborhoods, and GBK-GNN learns gated neighborhood kernels \citep{zhu2020beyond,pei2020geom,du2022gbk}. GCNII uses initial residuals and identity mapping for deeper propagation \citep{chen2020simple}. GPRGNN, FAGCN, ACM-GCN, and GAT-sep adapt propagation weights, frequency contributions, and propagation channels, respectively \citep{chien2021adaptive,bo2021beyond,luan2022revisiting, platonov2023critical}. Other approaches use global graph information, architecture search, or spectral filtering \citep{li2022finding,zheng2023auto,li2024pc,duan2024unifying}. CGNN uses commute-based aggregation, and L2DGCN combines graph enhancement with label selection \citep{pmlr-v267-zhuo25a,dingl2dgcn}. These approaches adapt how neighborhood information is selected and transformed. We analyze how sampled operations change the states used to compute adaptive weights at later layers.

\paragraph{Directionality and evaluation.}
MagNet represents edge direction with a magnetic Laplacian, and its signed-directed extension also incorporates polarity \citep{zhang2021magnet,ko2023spectral}. Direction-aware propagation uses edge orientation in message passing \citep{rossi2024edge}. In P3MP, the return path from an edge's target to its source determines when an earlier role can affect both endpoint states. Benchmark studies emphasize graph preprocessing, shared partitions, and the information available to each method \citep{platonov2023critical,dwivedi2023benchmarking,lim2021large}. We report these choices for both the feature-only and label-aware models.

\section{Implementation and Evaluation Protocol}\label{sec:implement}
\subsection{Architecture and execution}
By default, P3MP uses two signed layers. In the Cora depth comparison, we vary this stack while keeping the backbone and classifier unchanged.

\paragraph{Structural encoder and role decoder.}
We use two graph convolutional layers of width 128 for the structural encoder, with ReLU and dropout probability 0.5 after the first layer. The decoder concatenates the target and source embeddings, applies a linear map to width 128 and ReLU, and outputs three logits in subtractive, inactive, and additive order.

\paragraph{Prediction backbone.}
We use the backbone choices in Table \ref{tab:runner_datasets}. The graph attention backbone has two single-head layers of width 128, with ReLU and dropout after the first layer. We add a bias-free feature projection to the second-layer output before applying ReLU and dropout. On Chameleon and Squirrel, the backbone is an affine feature map followed by ReLU. The structural encoder uses the observed topology with either backbone.

For the Chameleon and Squirrel backbone, we apply inverted dropout to the activated feature map $H$ using independent Bernoulli entries:
\begin{equation}
\widetilde H=\frac{M\odot H}{q},\qquad M_{if}\sim\operatorname{Bernoulli}(q),\qquad q=\frac 1 4.
\end{equation}
Here, $\odot$ denotes elementwise multiplication, and $M$ is a mask with dropout probability $q$. All samples share the masked backbone output during training. At prediction, we use $H$.

\begin{table}[ht]
\centering
\caption{Prediction backbones for the eight primary benchmark graphs. Both choices use the observed graph in the structural encoder.}
\label{tab:runner_datasets}
\begin{tabular}{@{}ll@{}}
\toprule
Dataset & Prediction backbone \\
\midrule
Roman-Empire & Two-layer single-head GAT \\
Minesweeper & Two-layer single-head GAT \\
Amazon-Ratings & Two-layer single-head GAT \\
Chameleon & Feature MLP \\
Squirrel & Feature MLP \\
Actor & Two-layer single-head GAT \\
Cornell & Two-layer single-head GAT \\
Wisconsin & Two-layer single-head GAT \\
\bottomrule
\end{tabular}
\end{table}

\paragraph{Signed layers and classifier.}
Each signed layer has separate bias-free maps from width 128 to 64 for the target and source representations. We concatenate these projections, apply an affine map to width 128 and ReLU, and produce a scalar score with a final affine map. After thresholding the score as in Eq. \ref{eq:mass}, we sum the gated source values at their targets, project the aggregate affinely to width 128, and add a bias-free self-projection. ReLU and independent samplewise dropout with probability 0.5 follow the update. We initialize each layer's subtractive-strength parameter at 1.0 before softplus. A shared linear classifier maps the final hidden vectors to $C$ logits.

\paragraph{Optimization.}
We optimize with Adam, starting with a learning rate of $10^{-3}$ and weight decay $5\times10^{-4}$. We halve the learning rate when 50 consecutive validation evaluations have passed since the last improvement, down to a lower bound of $10^{-5}$. We train for at most 2000 epochs and stop after 200 consecutive epochs since the last strictly improving checkpoint. Both structural penalties apply from the first update. For target-class sample probabilities $P_{ik}$, we compute mixture NLL in the numerically stable form
\begin{equation}
\mathcal L_{\mathrm{cls},i}=\log K-\operatorname{logsumexp}_{k}(\log P_{ik}).
\end{equation}

Model training and prediction use 32-bit floating-point arithmetic for structural logits, edge scores, message weights, classifier logits, and probability calculations. Finite-state enumeration and local-gradient verification use double precision.

\subsection{Data splits, preprocessing, and model selection}
We use all observed features and edges for transductive propagation. Each of the ten main-benchmark partitions has one initialization, and all methods and input variants share those partitions. The disjoint training, validation, and test sets contain 10\%, 25\%, and 65\% of the nodes. On Cornell and Wisconsin, we use 20\%, 40\%, and 40\%.

We normalize nonzero feature rows to unit Euclidean norm and leave zero rows unchanged. We coalesce repeated directed edges, keeping their supplied directions, and assign a role distribution to each direction. The graph convolution and attention encoders use their standard self-loop handling. The signed stack uses the coalesced observed support and the learned self-contribution in Eq. \ref{eq:signed-aggregation}.

Within each split, we pair initialization and evaluation streams across compared configurations. Evaluation uses a separate random-number state and categorical role sampling with dropout disabled, leaving the training state unchanged. After each epoch, we evaluate validation ROC-AUC on Minesweeper and validation accuracy on the other graphs. We select a checkpoint whenever its validation score strictly improves, keeping the earlier checkpoint in a tie. Final test metrics use the selected parameters and averaged class probabilities.

We predict the class with the largest averaged probability. Test NLL is the mean of $-\log\hat p_{i,y_i}^{(K)}$ over test nodes, and Minesweeper ROC-AUC uses the averaged positive-class probability. Each run consists of a partition and initialization and serves as the statistical unit. We report means and sample standard deviations over runs. Appendix \ref{sec:app:controls} describes the procedure.

The larger-graph experiments use NVIDIA A100 GPUs with 80 GB per device. Shared encoders run once per forward computation, and the sample losses enter one joint backward pass. Appendix \ref{sec:app:memory} describes partitioned propagation and activation recomputation, including scorer and gate derivatives on edges with zero forward messages.

\subsection{Partitioned computation and activation storage}\label{sec:app:memory}
Graph data are distributed across $P$ devices with replicated model parameters. Device $p$ owns node set $\mathcal V_p$, and these sets partition $\mathcal V$. It evaluates incoming edges to its nodes in disjoint chunks $\mathcal C_{p,1},\ldots,\mathcal C_{p,J_p}$, where $J_p$ is its chunk count.

\paragraph{Layer computation.}
For sample $k$ and layer $\ell$, device $p$ obtains the source states needed by the current chunk, evaluates its endpoint scores and gates, and accumulates the messages at its owned targets. Writing $g_i^{(\ell,k)}$ for the aggregate before output projection gives
\begin{equation}
 g_i^{(\ell,k)}=
 \sum_{r=1}^{J_p}\sum_{(i,j)\in\mathcal C_{p,r}}
 u_{ij}^{(\ell,k)}d_{ij}(Z^{(k)})v_j^{(\ell,k)},
 \qquad i\in\mathcal V_p.
\end{equation}
The output map, self-contribution, and activation complete Eq. \ref{eq:signed-aggregation}. Partitioning the sum into chunks changes the accumulation order and leaves the mathematical operator unchanged. Shared-weight controls first average each edge's weights over samples and then apply the sample-specific gates and values.

\paragraph{Node activations and temporary edge values.}
Let $H_p^{(\ell,k)}$ collect the stored differentiable states of nodes in $\mathcal V_p$. The current chunk forms edge messages, endpoint concatenations, and scorer activations, then releases them after accumulation. Received source states remain available through their layer. Structural probabilities and sampling random-number states are maintained separately.

Activation checkpointing stores a subset of the node states and recomputes intermediate layers during the backward pass. Let $\mathcal Q\subseteq\{0,\ldots,L\}$ be the stored layer indices, including the initial and final states. For $K$ samples, the node-activation storage has order
\begin{equation}
 O\left(|\mathcal V_p|d_0+
 K\sum_{\ell\in\mathcal Q\setminus\{0\}}|\mathcal V_p|d_\ell\right),
\end{equation}
Here, $d_\ell$ is the width of layer $\ell$, and the initial state is shared across samples. We store the current input and output node states, the current edge chunk, and its required source states in GPU memory. Pinned host memory stores longer-lived node checkpoints and infrequently accessed features. The full input-feature collection is distributed across host processes. During differentiation, we reconstruct intermediate message vectors from node states.

\paragraph{Backward computation.}
During the backward pass, we recompute chunk messages and accumulate derivatives for both endpoint states, scorer parameters, and role logits. We use the same Gumbel variables and dropout masks as in the forward computation, so sampled roles and straight-through Jacobians remain unchanged, including those of inactive draws.

For a source node $j$, let $\operatorname{own}(j)$ denote its owning device. Each device groups contributions to the gradient of $h_j^{(\ell,k)}$ by $\operatorname{own}(j)$ and sends them to that device through an all-to-all exchange. The owner adds the received contributions to its local derivatives. An all-reduce sums gradients of the replicated parameters across devices before the optimizer step. Because loss normalization uses the global node and edge counts in Eq. \ref{eq:total}, the partitioned computation has the same full-graph objective and joint update over samples.

\subsection{Training algorithm}
Algorithm \ref{alg:p3mp} uses the straight-through signed gate $\widehat d_{ij}$ and activity gate $\widehat b_{ij}$ from Appendix \ref{sec:app:execution}. We denote the unnormalized weight-penalty sum by $S_{\mathrm{sp}}$.
\begin{algorithm}[ht]
\caption{P3MP training epoch}\label{alg:p3mp}
\begin{algorithmic}[1]
\REQUIRE $X,A_{\mathrm{obs}}$, training labels and their indicators $Y_{\mathrm{tr}}$, depth $L$, sample count $K$, model parameters $(\theta,\phi)$
\STATE Form $X_{\mathrm{st}}=X$ or $[X\|Y_{\mathrm{tr}}]$ for the selected input variant
\STATE Compute shared $H^{(0)}=B_\theta(X,A_{\mathrm{obs}})$ and structural probabilities $\{\pi_{ij}\}$
\STATE Compute $\mathcal L_{\mathrm{KL}}$ and $\mathcal L_{\mathrm{act}}$ from Eq. \ref{eq:struct-loss}
\STATE Set $S_{\mathrm{sp}}\gets0$
\FOR{$k=1,\ldots,K$}
\STATE Draw a hard straight-through role assignment $Z^{(k)}$ and use its role indicators at every layer
\STATE Set $H^{(0,k)}\gets H^{(0)}$
\FOR{$\ell=0,\ldots,L-1$}
\STATE Compute target and source projections, scores, and weights on all observed edges using Eq. \ref{eq:mass}
\STATE Apply Eq. \ref{eq:signed-aggregation} with $\widehat d_{ij}$, followed by ReLU and samplewise dropout
\STATE Add $\sum_{(i,j)\in\mathcal E_{\mathrm{obs}}}u_{ij}^{(\ell,k)}\widehat b_{ij}$ to $S_{\mathrm{sp}}$
\ENDFOR
\STATE Compute sample class probabilities with the shared classifier
\ENDFOR
\STATE Average sample probabilities and compute classification NLL on $\mathcal V_{\mathrm{tr}}$
\STATE Set $\mathcal L_{\mathrm{sparse}}\gets S_{\mathrm{sp}}/(nLK)$ and assemble Eq. \ref{eq:total}
\STATE Update $(\theta,\phi)$ with one backward pass and an Adam step
\end{algorithmic}
\end{algorithm}

\subsection{Operation count}
Let $d_{\mathrm{enc}}$, $d_{\mathrm{val}}$, and $d_{\mathrm{st}}$ denote the hidden, value, and structural-input widths, respectively. Structural encoding and pairwise decoding require
\begin{equation}
O(nd_{\mathrm{st}}d_{\mathrm{enc}}+nd_{\mathrm{enc}}^2+md_{\mathrm{enc}}^2).
\end{equation}
The edge term includes the decoder's hidden layer. The two-layer single-head GAT backbone requires $O(ndd_{\mathrm{enc}}+nd_{\mathrm{enc}}^2+md_{\mathrm{enc}})$ operations. The feature MLP requires $O(ndd_{\mathrm{enc}})$. For each signed layer, the cost is
\begin{equation}
O(nd_{\mathrm{enc}}^2+nd_{\mathrm{enc}}d_{\mathrm{val}}+md_{\mathrm{enc}}d_{\mathrm{val}}).
\end{equation}
These terms account for the self-projection, target, value, and output maps, as well as the scorer on observed edges. Combining shared encoding with the $K$ sample computations yields
\begin{equation}
\begin{split}
O\bigl(&n(d+d_{\mathrm{st}})d_{\mathrm{enc}}+nd_{\mathrm{enc}}^2+md_{\mathrm{enc}}^2\\
&+KL[nd_{\mathrm{enc}}^2+nd_{\mathrm{enc}}d_{\mathrm{val}}+md_{\mathrm{enc}}d_{\mathrm{val}}]+Knd_{\mathrm{enc}}C\bigr).
\end{split}
\label{eq:complexity}
\end{equation}
Equation \ref{eq:complexity} counts one forward evaluation, with shared encoding performed once and $m$ covering every observed edge, including zero-message edges. Training uses activation storage and recomputation, as described in Appendix \ref{sec:app:memory}. Backward recomputation adds chunk passes with the same linear dependence on graph size for fixed width, depth, class count, and sample count. Prediction disables gradient computation.

\section{Analytical Details}
\subsection{Sampling covariance and target-state feedback}\label{sec:app:feedback}
\paragraph{First-layer moments.}
All samples use the same weights and values before the first signed update. For independent edge roles with the moments in Eq. \ref{eq:gate-moments}, define the incoming aggregate as $G_i=\sum_{j\in\mathcal N_i(A_{\mathrm{obs}})}u_{ij}D_{ij}v_j$. Its moments are
\begin{equation}
\mathbb E[G_i]=\sum_j u_{ij}\mu_{ij}v_j,\qquad
\mathbb E\|G_i-\mathbb E[G_i]\|_2^2=\sum_j u_{ij}^2\nu_{ij}\|v_j\|_2^2.
\end{equation}
Each source contributes to the aggregate variance in proportion to its squared weight and gate variance. The second-layer scorer uses the node states produced by these sampled aggregates.

\begin{proposition}[Target-state feedback]\label{prop:target-feedback}
Consider the coupled two-layer comparison of Section \ref{sec:theory}, with a factorized role law, positive layerwise subtractive strengths, and distinct endpoints on every incoming sampled edge of node $i$. The expected difference in preactivation between persistent and resampled roles, projected along $w$, equals $\mathcal F_i(w)$ in Eq. \ref{eq:direct-feedback}. If $\eta_e(U_e^t-U_e^s)\ge0$ almost surely for every positive-probability ordered role pair, then $\mathcal F_i(w)\ge0$. A strict inequality on a set of positive probability for at least one pair gives strict positivity.

Let $g_e=W_o^{(0)}u_e^{(0)}v_j^{(0)}$ be the first-layer message before gating, $C_e$ a target-state Lipschitz constant of the second-layer weight along its role interventions, and $\nu_e=\operatorname{Var}(D_e)$. Then
\begin{equation}
|\mathcal F_i(w)|\le\chi\sum_{e=(i,j)}\nu_e\,
\mathbb E_{Z_{-e}}[|\eta_e|C_e\|g_e\|_2],\qquad
\chi=\max\left\{1,\frac{\gamma^{(0)}}{\gamma^{(1)}}\right\}.
\label{eq:feedback-bound}
\end{equation}
Here, $\chi$ accounts for the layers' different subtractive strengths.
\end{proposition}

Writing $U_e=U_e^{z_e}$, the edge message satisfies the standard covariance identity
\begin{equation}
\mathbb E[U_eD_eV_e]=\mu_e\mathbb E[U_eV_e]+\operatorname{Cov}(U_eV_e,D_e).
\label{eq:persistent-message}
\end{equation}
The covariance is coordinate-wise for vector-valued messages.

For an independent second-layer gate $D'_e$ with the same marginal law as $D_e$,
\begin{equation}
\mathbb E[U_eD_eV_e]-\mathbb E[U_eD'_eV_e]
=\operatorname{Cov}(U_eV_e,D_e).
\end{equation}
Independence of $D'_e$ implies $\mathbb E[U_eD'_eV_e]=\mu_e\mathbb E[U_eV_e]$, where $\mu_e=\mathbb E[D_e]$. For a constant weight, the contrast reduces to $U_e\operatorname{Cov}(V_e,D_e)$, so it arises entirely through the source state.

\paragraph{Distinct-endpoint edges.}
For a distinct-endpoint edge, the source value $V_e$ is independent of its own edge role conditional on $Z_{-e}$ because the first-layer update is simultaneous. The factorized role law also gives $\mathbb E[D_e\mid Z_{-e}]=\mu_e$. It follows that
\begin{equation}
\mathbb E[U_e(D_e-\mu_e)V_e\mid Z_{-e}]
=V_e\operatorname{Cov}(U_e,D_e\mid Z_{-e}).
\end{equation}
For the three categorical states, symmetrization yields
\begin{equation}
\operatorname{Cov}(U_e,D_e\mid Z_{-e})
=\sum_{s<t}\pi_e^s\pi_e^t(U_e^t-U_e^s)(d_t^{(1)}-d_s^{(1)}).
\end{equation}
The same symmetrization holds for any finite collection of gate values. For a binary role with probability $p$ for state 1, let $U_0,U_1$ be the two weights and $d_0,d_1$ the two gate values. Then
\begin{equation}
\operatorname{Cov}(U,D\mid Z_{-e})
=p(1-p)(U_1-U_0)(d_1-d_0).
\end{equation}
A binary activity gate uses $(d_0,d_1)=(0,1)$, and a binary signed gate uses $(-\gamma,1)$. P3MP's tri-state covariance contains the pairs $(-1,0)$, $(0,+1)$, and $(-1,+1)$. The first two change activity; the third changes polarity.

We obtain Eq. \ref{eq:direct-feedback} by applying $w^\top W_o^{(1)}$ and summing over incoming edges. The two second-layer computations use the same first-layer state in the learned self path, so the self-contribution and common bias cancel. The first-layer self map still contributes to the states from which we compute $U_e^s$.

\paragraph{The finite-sample sharing contrast.}
Expanding $\bar U\bar A$ in Eq. \ref{eq:sample-sharing-identity} produces $K$ same-sample terms with expectation $\mathbb E[UA]$ and $K(K-1)$ cross-sample terms with expectation $\mathbb E[U]\mathbb E[A]$. Therefore,
\begin{equation}
\mathbb E\left[\frac1K\sum_kU_kA_k-\bar U\bar A\right]
=\left(1-\frac1K\right)\operatorname{Cov}(U,A).
\label{eq:mass-sharing-covariance}
\end{equation}

During training, let $\Omega$ collect the shared backbone and structural-encoder dropout. The samples are independent conditional on $\Omega$, so the expectation becomes
\begin{equation}
\left(1-\frac1K\right)\mathbb E_\Omega[\operatorname{Cov}(U,A\mid\Omega)].
\end{equation}
The same calculation for the activity indicator $I_k=\mathbb I(z^{(k)}\ne0)$ yields the weight-penalty contrast $(1-1/K)\operatorname{Cov}(U,I)$. Thus, weight sharing also affects the penalty on active edges.

For sample $k$, write $U_{e,k}$, $V_{e,k}$, $D_{e,k}$, and $D'_{e,k}$ for the weight, value, persistent gate, and resampled gate, respectively. With $\bar U_e=K^{-1}\sum_k U_{e,k}$, the four edge contributions before output projection are
\begin{equation}
\begin{aligned}
M_e^{\mathrm{p,l}}&=K^{-1}\sum_k U_{e,k}D_{e,k}V_{e,k},&
M_e^{\mathrm{p,s}}&=K^{-1}\sum_k \bar U_eD_{e,k}V_{e,k},\\
M_e^{\mathrm{r,l}}&=K^{-1}\sum_k U_{e,k}D'_{e,k}V_{e,k},&
M_e^{\mathrm{r,s}}&=K^{-1}\sum_k \bar U_eD'_{e,k}V_{e,k}.
\end{aligned}
\end{equation}
Applying the sample-covariance identity in Eq. \ref{eq:mass-sharing-covariance} to each sampling rule yields
\begin{equation}
\begin{split}
&\mathbb E \left[(M_e^{\mathrm{p,l}}-M_e^{\mathrm{p,s}})
-(M_e^{\mathrm{r,l}}-M_e^{\mathrm{r,s}})\right]\\
&\qquad=\left(1-\frac1K\right)
\mathbb E[(U_e-\mathbb EU_e)(D_e-\mu_e)V_e].
\end{split}
\label{eq:general-route-interaction}
\end{equation}
For distinct endpoints, $\mathbb E[(D_e-\mu_e)V_e]=0$, leaving $(1-1/K)\operatorname{Cov}(U_eV_e,D_e)$. Local and shared weights coincide at $K=1$.

\paragraph{Feedback under shared weights.}
Under the distinct-endpoint assumption, $V_{e,k}$ is independent of its own edge gate $D_{e,k}$. With shared weights, the expected difference between persistent and resampled contributions is
\begin{equation}
\begin{aligned}
\mathbb E[M_e^{\mathrm{p,s}}-M_e^{\mathrm{r,s}}]
&=\frac1{K^2}\sum_{k,b}
\mathbb E[U_{e,b}(D_{e,k}-D'_{e,k})V_{e,k}]\\
&=\frac1K\operatorname{Cov}(U_eV_e,D_e).
\end{aligned}
\end{equation}
The cross-sample terms are zero because $U_{e,b}$ is independent of sample $k$ and $\mathbb E[(D_{e,k}-D'_{e,k})V_{e,k}]=0$. The same-sample terms contribute to the role-state covariance. Projection and summation over the contrasts establish Eq. \ref{eq:shared-feedback}.

\paragraph{The two weighting contrasts.}
For distinct endpoints, independence of $V_e$ and $D_e$ implies $\mathbb E[D_eV_e]=\mu_e\mathbb E[V_e]$. Expanding the two covariances yields
\begin{equation}
\operatorname{Cov}(U_e,D_eV_e)
=\operatorname{Cov}(U_eV_e,D_e)+\mu_e\operatorname{Cov}(U_e,V_e).
\end{equation}
Under resampling, the covariance is $\mu_e\operatorname{Cov}(U_e,V_e)$. Applying Eq. \ref{eq:mass-sharing-covariance}, projecting, and summing proves Eq. \ref{eq:allocation-decomposition}.

\begin{corollary}[Source-only zero case]\label{cor:source-only}
Under Proposition \ref{prop:target-feedback}, suppose each second-layer weight depends only on its source representation. Then $\mathcal F_i(w)=\mathcal J_i(w)=0$ for every $w$ and $K$.
\end{corollary}
\begin{proof}
Changing $z_e$ leaves the first-layer source representation of a distinct-endpoint edge unchanged. For every role pair, $U_e^t=U_e^s$, and substitution into Eqs. \ref{eq:direct-feedback} and \ref{eq:graph-interaction} proves the claim. Constant weights and weights computed from $H^{(0)}$ also satisfy this condition.
\end{proof}

\paragraph{Resampled preactivation dispersion.}
For constant projected source values and mean-zero independently resampled gates, let $\upsilon_k$ denote the projected self-contribution and bias in sample $k$. Conditional on the first-layer states, both weighting rules have mean $\upsilon_k$ because the resampled gates have mean zero. Their conditional variances are $\sum_e\nu_e\eta_e^2U_{e,k}^2$ and $\sum_e\nu_e\eta_e^2\bar U_e^2$. The law of total variance then yields
\begin{equation}
\begin{aligned}
\operatorname{Var}(w^\top a_i^{\mathrm{r,l},k})
&=\operatorname{Var}(\upsilon_k)+\sum_e\nu_e\eta_e^2\mathbb E[U_e^2],\\
\operatorname{Var}(w^\top a_i^{\mathrm{r,s},k})
&=\operatorname{Var}(\upsilon_k)+\sum_e\nu_e\eta_e^2\mathbb E[\bar U_e^2].
\end{aligned}
\end{equation}
Independent first-layer histories imply $\operatorname{Var}(\bar U_e)=\operatorname{Var}(U_e)/K$. The mean weights agree, so subtracting the variances yields
\begin{equation}
\operatorname{Var}(w^\top a_i^{\mathrm{r,l},k})-
\operatorname{Var}(w^\top a_i^{\mathrm{r,s},k})
=\left(1-\frac1K\right)\sum_{e=(i,j)}\nu_e\eta_e^2\operatorname{Var}(U_e).
\end{equation}
This variance difference applies to the constant-source, mean-zero construction in Appendix \ref{sec:app:feature_coupling_design}.

\paragraph{Weight response along an edge intervention.}
Let $a_{i,-e}$ be the first-layer preactivation with edge $e$'s message removed, including the self-contribution, bias, and other incoming messages. If $g_e=W_o^{(0)}u_e^{(0)}v_j^{(0)}$ is the message before role gating, we can write the state at a continuous gate value $t$ as
\begin{equation}
h_i^{(1)}(t)=\operatorname{ReLU}(a_{i,-e}+tg_e),
\qquad -\gamma^{(0)}\le t\le1.
\end{equation}
We evaluate the second-layer weight at this target state and the unchanged source state to obtain $U_e(t)$. The scorer, absolute value, and threshold make $U_e(t)$ continuous and piecewise affine. Writing $J_i(t)$ for the diagonal derivative of the first-layer ReLU, the chain rule yields
\begin{equation}
\frac{\mathrm dU_e(t)}{\mathrm dt}
=\bigl(\nabla_{h_i^{(1)}}u_e^{(1)}\bigr)^\top J_i(t)g_e
\quad\text{almost everywhere}.
\end{equation}
Integrating between the two gate values yields the weight difference $U_e^t-U_e^s$ in Eq. \ref{eq:direct-feedback}. Its sign depends on the scorer gradient and the transmitted message. Below the weight threshold, the derivative is zero. In the ReLU enumeration, we evaluate both response directions and the zero-response case.

\paragraph{Feedback magnitude and threshold support.}
For the intervention path $h_i^{(1)}(t)$ above, ReLU's nonexpansiveness gives
\begin{equation}
\|h_i^{(1)}(d_t^{(0)})-h_i^{(1)}(d_s^{(0)})\|_2
\le |d_t^{(0)}-d_s^{(0)}|\,\|g_e\|_2.
\end{equation}
The finite two-layer scorer is Lipschitz; absolute value and positive-part thresholding are each 1-Lipschitz. If $C_e$ is a target-state Lipschitz constant of the weight, then
\begin{equation}
|U_e^t-U_e^s|\le C_e\|g_e\|_2|d_t^{(0)}-d_s^{(0)}|.
\end{equation}
For any given $Z_{-e}$, we can choose $C_e$ uniformly over the three role interventions. For the ordered pairs $(-1,0)$, $(0,+1)$, and $(-1,+1)$, the ratios of first-layer to second-layer gate differences are
\begin{equation}
\frac{\gamma^{(0)}}{\gamma^{(1)}},\qquad 1,\qquad
\frac{1+\gamma^{(0)}}{1+\gamma^{(1)}}.
\end{equation}
Each ratio is bounded by $\chi$ from Eq. \ref{eq:feedback-bound}. The triangle inequality and the categorical variance identity
\begin{equation}
\sum_{s<t}\pi_e^s\pi_e^t(d_t^{(1)}-d_s^{(1)})^2
=\operatorname{Var}(D_e)=\nu_e
\end{equation}
complete the proof of Eq. \ref{eq:feedback-bound}. The sign condition follows because the gate differences are positive for ordered role pairs. An edge's target feedback is zero when its first-layer weight is zero or when every second-layer role intervention gives a score within the threshold.

\paragraph{Self-edges and greater depth.}
For an observed self-edge $e=(i,i)$, a role intervention can change both endpoint states. The direct sampling contrast remains $\operatorname{Cov}(U_eV_e,D_e)$, and its sharing contrast remains Eq. \ref{eq:general-route-interaction}. Their difference includes the source-state term:
\begin{equation}
\mathbb E[(U_e-\mathbb EU_e)(D_e-\mu_e)V_e]
=\operatorname{Cov}(U_eV_e,D_e)
-\mathbb E[U_e]\operatorname{Cov}(V_e,D_e).
\end{equation}
We add these contributions to the distinct-endpoint sums. The learned self path and a sampled observed self-edge contribute separately to the update.

\paragraph{The role of directed return paths.}
At depth $L$, we first generate each sample's preceding $L-1$ layers with persistent roles and local weights. At the final layer, we compare keeping the original gate with independently resampling it, and local weights with shared weights. All four computations start from the same preceding states. For one edge, define
\begin{equation}
\Gamma_e=\operatorname{Cov}(U_eV_e,D_e)
-\mathbb E[U_e]\operatorname{Cov}(V_e,D_e).
\end{equation}
Here, $\Gamma_e$ is the edge's contribution to the final-layer interaction, and $U_e$, $V_e$, and $D_e$ are the weight, value, and gate at that layer. Expanding the sample-mean products as in Eq. \ref{eq:general-route-interaction} yields $(1-1/K)\Gamma_e$ at any depth. We project with $w^\top W_o^{(L-1)}$ and sum over incoming edges to obtain the node-level interaction.

\begin{proposition}[Return-path criterion]\label{prop:return-path}
Let $d_e$ be the shortest directed distance from the target of edge $e=(i,j)$ to its source, with $d_e=\infty$ when there is no directed return path. In P3MP's sample-specific signed stack, the state used for the layer-$L$ source value is independent of $z_e$, conditional on $Z_{-e}$, whenever $L<d_e+2$. Under this condition, $\operatorname{Cov}(V_e,D_e)=0$ and
\begin{equation}
\Gamma_e=\mathbb E_{Z_{-e}}\left[
 V_e\sum_{s<t}\pi_e^s\pi_e^t(d_t^{(L-1)}-d_s^{(L-1)})(U_e^t-U_e^s)\right].
\label{eq:return-path-feedback}
\end{equation}
To obtain $U_e^s$, we hold the role on $e$ at $s$ throughout the preceding layers. Under the return-path condition, source-only weights give $\Gamma_e=0$. In a directed acyclic graph, the condition holds for every edge at every finite depth.
\end{proposition}

\begin{proof}
Changing $z_e$ first affects target $i$ at the first signed layer. After $\ell$ layers, each affected node is reachable from $i$ by a directed path of length at most $\ell-1$. We prove this by induction. The self map and target argument can carry the dependence forward at an affected node, and a source-dependent message can transfer it along one observed edge. Reaching source $j$ by layer $L-1$ requires $d_e\le L-2$.

When $L<d_e+2$, the source value $V_e$ stays unchanged under every role intervention. The factorized role law gives independence from $D_e$, and the categorical covariance formula yields Eq. \ref{eq:return-path-feedback}. A source-only weight also stays unchanged. Any directed return path would complete a cycle with $j\to i$, so $d_e=\infty$ for every edge in an acyclic graph.
\end{proof}

When an edge intervention changes both endpoints, write $U_e^s,V_e^s$ for the final weight and value obtained by using role $s$ throughout the preceding layers. The correction has the pairwise form
\begin{equation}
\begin{split}
\Gamma_e=\mathbb E_{Z_{-e}}\sum_{s<t}\pi_e^s\pi_e^t
(d_t^{(L-1)}-d_s^{(L-1)})\Bigg[
&\frac{V_e^t+V_e^s}{2}(U_e^t-U_e^s)\\
+&\left(\frac{U_e^t+U_e^s}{2}-\mathbb E[U_e]\right)(V_e^t-V_e^s)
\Bigg].
\end{split}
\end{equation}
Symmetrizing the covariances and expanding $U_e^tV_e^t-U_e^sV_e^s$ separates the weight response from the source response. For a self-edge, $d_e=0$, so both responses can contribute from the second signed layer. The directed distance determines the earliest layer at which the source-response term can contribute.

\subsection{Sample-weight permutation identity}\label{sec:app:permutation}
Let $K\ge2$ and let $\mathfrak D_K$ be the set of permutations of $\{1,\ldots,K\}$ with no fixed points. Choose $\sigma$ uniformly from $\mathfrak D_K$, independently of the sampled representations and gates. The same permutation is applied to the weight vector over all observed edges. For a given edge, write $A_k=D_kV_k$ for its gated source value and define
\begin{equation}
M_{\mathrm l}=\frac1K\sum_k U_kA_k,\qquad
M_{\mathrm s}=\bar U\bar A,\qquad
M_\sigma=\frac1K\sum_k U_{\sigma(k)}A_k.
\end{equation}
These are the local, shared, and permuted sample-average messages before output projection. Permuting complete weight vectors leaves their joint empirical distribution and all empirical moments unchanged.

For any sample $k$, symmetry gives $\Pr(\sigma(k)=b)=1/(K-1)$ for each $b\ne k$. Conditional on the sampled batch,
\begin{equation}
\begin{aligned}
\mathbb E_\sigma[M_\sigma]
&=\frac1{K(K-1)}\sum_k\sum_{b\ne k}U_bA_k\\
&=\frac{K\bar U\bar A-M_{\mathrm l}}{K-1}.
\end{aligned}
\end{equation}
Subtracting from $M_{\mathrm l}$, applying the common output projection, and summing over incoming edges proves Eq. \ref{eq:permutation-batch}. The learned self-contribution is identical in the three computations and cancels from the contrast.

For independent, identically distributed pairs $(U_k,A_k)$, every cross-sample expectation factors as $\mathbb E[U]\mathbb E[A]$. Therefore,
\begin{equation}
\mathbb E[M_{\mathrm l}-M_\sigma]=\operatorname{Cov}(U,A).
\end{equation}
Under the two-layer, distinct-endpoint comparison, using persistent and independently resampled gate-value products gives
\begin{equation}
\begin{aligned}
\mathbb E[w^\top(\bar a_i^{\mathrm{r,l}}-\bar a_i^{\mathrm{r},\sigma})]
&=\mathcal S_i(w),\\
\mathbb E[w^\top(\bar a_i^{\mathrm{p,l}}-\bar a_i^{\mathrm{p},\sigma})]
&=\mathcal S_i(w)+\mathcal F_i(w).
\end{aligned}
\end{equation}
Their difference is the permutation interaction $\mathcal J_i^\sigma(w)$, which satisfies
\begin{equation}
\mathcal J_i^\sigma(w)=\mathcal F_i(w)
=\frac K{K-1}\mathcal J_i(w).
\end{equation}
The expectation includes both role draws and the independent permutation. In the three-node construction of Appendix \ref{sec:app:feature_coupling_design}, we obtain $\mathcal J_2^\sigma(1)=\rho/4$. A shared mean contains each sample's own weight. Under permutation, each sample receives another sample's weight, preserving its original magnitude.

\subsection{Persistence across layers: an auxiliary example}
In the self-edge construction below, identical marginal role probabilities yield different input-response orders. The constant-weight case isolates the contribution of cross-layer dependence.

Consider one node with an observed self-edge, two signed layers, zero self maps, and encoded input $x\in[0,1]$. All scorer, projection, bias, and classifier parameters are finite and independent of the activity probability $\alpha\in(0,1)$. In subtractive, inactive, and additive order, the edge probabilities are $(\alpha/2,1-\alpha,\alpha/2)$. Let $F_\tau(x,\alpha)$ be the population positive-class probability for $\tau\in\{\mathrm p,\mathrm r,\mathrm{mean}\}$, denoting persistent sampling, layerwise resampling, and expected signed gates. Define $R_\tau(x,\alpha)=F_\tau(x,\alpha)-F_\tau(0,\alpha)$ as the response to changing the encoded input from zero to $x$.

\begin{theorem}[Activity-order separation]
Independently resampled roles and expected signed gates satisfy
\begin{equation}
|R_{\mathrm r}(x,\alpha)|\le \alpha^2,\qquad R_{\mathrm{mean}}(x,\alpha)=O(\alpha^2)
\quad\text{as }\alpha\downarrow0.
\end{equation}
Persistent sampling with thresholded weight $u(h)=h$ realizes
\begin{equation}
R_{\mathrm p}(x,\alpha)=\frac \alpha2\left(\operatorname{sigmoid}(x^4)-\frac12\right).
\label{eq:activity-separation}
\end{equation}
For $x>0$, this response depends on activity to first order.
\end{theorem}

\begin{proof}
Couple the resampled predictions at $x$ and $0$ using identical layerwise roles. An inactive role at either layer makes the resulting states agree because the self maps are zero. The two sample probabilities can differ only when both roles are active, which has probability $\alpha^2$.

For expected gates, write the layer-$\ell$ gate as $\alpha\chi_\ell$, with $\chi_\ell=(1-\gamma^{(\ell)})/2$. If $g_\ell(h)$ is its message map before gating, the recurrence is
\begin{equation}
h^{(\ell+1)}(x,\alpha)=\operatorname{ReLU}(\alpha\chi_\ell g_\ell(h^{(\ell)}(x,\alpha))+b^{(\ell)}).
\end{equation}
Each message map $g_\ell$ is locally Lipschitz because the scorer and projections have finite parameters. The first-layer state difference is $O(\alpha)$, and both states lie in a compact set as $\alpha$ approaches zero. The second message map contributes another factor $\alpha$, so the state difference is $O(\alpha^2)$. Applying the finite softmax classifier leaves this order unchanged.

For persistence, choose unit scalar target, value, and output maps, zero biases, and $\gamma=1$. The scorer $\psi(h,h)=\operatorname{ReLU}(h)+\lambda_{\mathrm{sc}}$ gives $u(h)=h$ on nonnegative states at the original positive threshold. An additive draw yields $x\to x^2\to x^4$. Either the other role yields zero. With binary logits $(0,h^{(2)})$, we obtain Eq. \ref{eq:activity-separation}. We embed these scalar maps in the stated hidden widths through one coordinate. Dividing by $\alpha$ gives a positive limit for the persistent response when $x>0$, and both comparison responses converge to zero.
\end{proof}

\paragraph{Constant-weight calculation.}
Keeping the maps and readout above, we set $\psi(h,h)=1+\lambda_{\mathrm{sc}}$ so that $u\equiv1$. Direct propagation yields
\begin{equation}
\begin{aligned}
R_{\mathrm p}^{\mathrm{const}}(x,\alpha)&=\frac \alpha2\left(\operatorname{sigmoid}(x)-\frac12\right),\\
R_{\mathrm r}^{\mathrm{const}}(x,\alpha)&=\frac{\alpha^2}{4}\left(\operatorname{sigmoid}(x)-\frac12\right),\qquad
R_{\mathrm{mean}}^{\mathrm{const}}(x,\alpha)=0.
\end{aligned}
\end{equation}
With constant weights, the sampling rules yield different activity orders because both layers use a persistent active role.

\subsection{Straight-through execution}\label{sec:app:execution}
For one observed edge, let $\beta_{ij}^s$ be the structural logit of state $s\in\{-1,0,+1\}$ and let $\xi_{ij}^s$ be independent standard Gumbel variables. The relaxed weights and hard indicators are
\begin{equation}
w_{ij}^s=\frac{\exp((\beta_{ij}^s+\xi_{ij}^s)/T)}{\sum_t\exp((\beta_{ij}^t+\xi_{ij}^t)/T)},\qquad
b_{ij}^s=\mathbb I\left(s=\arg\max_t(\beta_{ij}^t+\xi_{ij}^t)\right).
\end{equation}
We use temperature $T>0$ for the relaxation. The hard draw follows the categorical probabilities defined by the logits. The straight-through indicators are
\begin{equation}
\widehat b_{ij}^s=\operatorname{stopgrad}(b_{ij}^s-w_{ij}^s)+w_{ij}^s,\qquad
\widehat d_{ij}=\widehat b_{ij}^+-\gamma\widehat b_{ij}^-,\qquad
\widehat b_{ij}=\widehat b_{ij}^++\widehat b_{ij}^-.
\end{equation}
Here, $\operatorname{stopgrad}$ acts as the identity in the forward computation and has zero derivative. We use hard role indicators in the forward pass and the relaxed softmax Jacobian in the backward pass. The signed gate $\widehat d_{ij}$ enters propagation, and the activity gate $\widehat b_{ij}$ enters the weight penalty.

For a message $M_{ij}=u_{ij}\widehat d_{ij}v_j$, the structural-logit derivative can be nonzero even when the hard role is inactive, and the forward message is zero. Backpropagation also passes through the preceding states used by the scorer and source projection. We evaluate gates on the full observed support to keep these derivative paths.

\paragraph{Local gradient verification.}
We compare computation on the complete observed support with an active-edge forward computation whose backward rule differentiates every observed gate. We use first-layer weights, logits $(-0.3,0.2,0.1)$, and temperature $0.5$. Increasing one selected logit coordinate by 2 covers all nine hard assignments on two edges. In double-precision CPU evaluation, the forward values and gradients agree for structural logits, input features, and scorer parameters. The inactive edge in the inactive-positive assignment of Table \ref{tab:inactive-gradient} has a nonzero straight-through derivative, so its role probabilities can change during training.

\begin{table}[ht]
\centering
\caption{Inactive-edge gradient in the two-edge verification. The hard forward message is zero in both cases. Differentiating every observed gate keeps the structural gradient of an inactive draw.}
\label{tab:inactive-gradient}
\begin{tabular}{@{}lc@{}}
\toprule
Differentiated gates & Structural gradient norm on the inactive edge \\
\midrule
All observed gates, straight-through & $0.0557$ \\
Active gates only & $0$ \\
\bottomrule
\end{tabular}
\end{table}

\subsection{Mixture loss and sample responsibilities}\label{sec:app:mixture}
The objective gap in Eq. \ref{eq:objective-gap} vanishes exactly when all samples assign the same target-class probability.

For a training node and a batch of role assignments, differentiation of Eq. \ref{eq:responsibilities} yields
\begin{equation}
\nabla_\theta\mathcal L_{\mathrm{cls},i}
=-\sum_{k=1}^Kr_{ik}\nabla_\theta\log P_{ik}.
\end{equation}
The responsibilities also follow from a variational identity. Let $\Delta_K$ be the probability simplex over $K$ samples. For any $\boldsymbol w\in\Delta_K$,
\begin{equation}
\sum_kw_k(-\log P_{ik})+\mathrm{KL}(\boldsymbol w\|\boldsymbol u_K)
=\mathcal L_{\mathrm{cls},i}+\mathrm{KL}(\boldsymbol w\|\boldsymbol r_i).
\end{equation}
Since every target probability is positive, the responsibility vector $\boldsymbol r_i$ uniquely minimizes the left side. Setting $\boldsymbol w=\boldsymbol u_K$ recovers the objective gap in Eq. \ref{eq:objective-gap}.

Let $Z^{(1:K)}=(Z^{(1)},\ldots,Z^{(K)})$ be an ordered batch drawn from $q_\phi^{\otimes K}$, the product of $K$ independent role laws. The expected categorical classification objective is
\begin{equation}
J_K(\theta,\phi)=\mathbb E\left[\frac1{|\mathcal V_{\mathrm{tr}}|}\sum_{i\in\mathcal V_{\mathrm{tr}}}\mathcal L_{\mathrm{cls},i}(Z^{(1:K)})\right].
\end{equation}
At a fixed dropout realization and fixed predictor parameters, its categorical structural derivative is
\begin{equation}
\nabla_\phi J_K=
\mathbb E\left[\left(\frac1{|\mathcal V_{\mathrm{tr}}|}\sum_{i\in\mathcal V_{\mathrm{tr}}}\mathcal L_{\mathrm{cls},i}\right)
\sum_{k=1}^K\nabla_\phi\log q_\phi(Z^{(k)})\right].
\end{equation}
This is the score-function derivative of the categorical objective. Training uses the straight-through surrogate from Appendix \ref{sec:app:execution}, with hard categorical forward draws and gate Jacobians from the relaxation.

\subsection{Thresholding and message-weight bounds}
For an observed edge, the probability of a nonzero message coefficient is
\begin{equation}
\begin{split}
\Pr(u_{ij}d_{ij}\ne0)={}&\pi_{ij}^+\Pr(|s_{ij}|>\lambda_{\mathrm{sc}}\mid z_{ij}=+1)\\
&+\pi_{ij}^-\Pr(|s_{ij}|>\lambda_{\mathrm{sc}}\mid z_{ij}=-1).
\end{split}
\end{equation}
An edge transmits a message when its role is active, and its score exceeds the threshold. Because the scorer uses intermediate states, an active edge can transmit at one layer and have zero weight at another.

For a subset $\mathcal T_i$ of the incoming neighbors, define its aggregate message and maximum source-value norm by
\begin{equation}
m_i=\sum_{j\in \mathcal T_i}u_{ij}d_{ij}v_j,\qquad
V_{i,\mathcal T}=\max_{j\in \mathcal T_i}\|v_j\|_2.
\end{equation}
We set the maximum to zero for an empty subset. By the triangle inequality,
\begin{equation}
\|m_i\|_2\le V_{i,\mathcal T}\sum_{j\in \mathcal T_i}u_{ij}|d_{ij}|
\le V_{i,\mathcal T}\max\{1,\gamma\}\sum_{j\in \mathcal T_i}u_{ij}b_{ij}.
\end{equation}
If value norms are bounded by $V_{\max}$ and layerwise channel strengths by $\gamma_{\max}$, averaging over nodes, layers, and role draws yields
\begin{equation}
\frac1{nL}\sum_{i,\ell}\mathbb E\|m_i^{(\ell)}\|_2
\le V_{\max}\max\{1,\gamma_{\max}\}\,
\mathbb E\mathcal L_{\mathrm{sparse}}.
\end{equation}
The bound applies to any designated edge subset. Replacing its weights with another nonnegative vector changes the aggregate by at most the weights' $\ell_1$ difference multiplied by the source-value scale and subtractive-strength factor. The output projection contributes a further factor $\|W_o\|_2$.

The transformations $u_{ij}\mapsto u_{ij}/c$ and $v_j\mapsto cv_j$, for $c>0$, preserve $u_{ij}d_{ij}v_j$, making the value scale relevant to the weight bound. In P3MP, the projected value also enters the scorer, so the learned parameters couple the two scales.

\subsection{Finite-margin identity for predictive probabilities}\label{sec:app:margin}
For two sets of finite logits labeled $\mathrm s$ and $\mathrm l$, define their target-versus-$c$ margin change in sample $k$ as
\begin{equation}
\delta_{i,c}^k=(o_{i,y_i}^{\mathrm l,k}-o_{i,c}^{\mathrm l,k})
-(o_{i,y_i}^{\mathrm s,k}-o_{i,c}^{\mathrm s,k}).
\end{equation}
Here, $o_{i,c}^{\kappa,k}$ is the logit for class $c$ in configuration $\kappa$. Let $p_{i,c}^{\mathrm s,k}$ denote the initial class probabilities, $P_{ik}^{\mathrm s}=p_{i,y_i}^{\mathrm s,k}$ their target coordinate, and $r_{ik}^{\mathrm s}=P_{ik}^{\mathrm s}/\sum_bP_{ib}^{\mathrm s}$ the responsibility for that sample.

\begin{proposition}[Finite-margin identity]
For a predictive average of $K$ samples, the two mixture losses satisfy
\begin{equation}
\mathcal L_i^{\mathrm s}-\mathcal L_i^{\mathrm l}
=\log\sum_{k=1}^K
\frac{r_{ik}^{\mathrm s}}
{P_{ik}^{\mathrm s}+\sum_{c\ne y_i}p_{i,c}^{\mathrm s,k}e^{-\delta_{i,c}^k}}.
\label{eq:finite-mixture-criterion}
\end{equation}
For $C\ge2$, let $\delta_{ik}^- =\min_{c\ne y_i}\delta_{i,c}^k$ and $\delta_{ik}^+=\max_{c\ne y_i}\delta_{i,c}^k$ be the smallest and largest margin changes. With
\begin{equation}
 g_{ik}(d)=\left[P_{ik}^{\mathrm s}+(1-P_{ik}^{\mathrm s})e^{-d}\right]^{-1},
\end{equation}
the function $g_{ik}$ is the target-probability ratio for a common margin increase $d$, and
\begin{equation}
\log\sum_k r_{ik}^{\mathrm s}g_{ik}(\delta_{ik}^-)
\le\mathcal L_i^{\mathrm s}-\mathcal L_i^{\mathrm l}
\le\log\sum_k r_{ik}^{\mathrm s}g_{ik}(\delta_{ik}^+).
\label{eq:branchwise-margin-bounds}
\end{equation}
\end{proposition}
\begin{proof}
From the softmax odds,
\begin{equation}
\frac{P_{ik}^{\mathrm l}}{P_{ik}^{\mathrm s}}
=\left[P_{ik}^{\mathrm s}+\sum_{c\ne y_i}p_{i,c}^{\mathrm s,k}e^{-\delta_{i,c}^k}\right]^{-1}.
\end{equation}
The ratio of averaged target probabilities is the responsibility-weighted mean of these sample ratios. Taking its logarithm proves Eq. \ref{eq:finite-mixture-criterion}. Bounding each competing margin by its samplewise minimum and maximum proves Eq. \ref{eq:branchwise-margin-bounds}. All denominators are positive because logits are finite. We verify the inequalities numerically in Table \ref{tab:theory-verification}.
\end{proof}

For the local and shared weighting comparison, the margin change includes the activation and classifier:
\begin{equation}
\delta_{i,c}^k=(w_{y_i}-w_c)^\top
[\operatorname{ReLU}(a_i^{k,\mathrm l})-\operatorname{ReLU}(a_i^{k,\mathrm s})].
\end{equation}
Here, $w_c^\top$ is row $c$ of the shared classifier. We evaluate this identity from samplewise logits. The interaction $\mathcal J_i(w)$ in Eq. \ref{eq:graph-interaction} is the expectation of a projected preactivation contrast.

\subsection{Standard Monte Carlo integration results}\label{sec:app:mc}
At prediction, P3MP's shared encoders are deterministic, and each sample draws an independent categorical role. For a selected predictor and node $i$, let $P_k=p_{i,y_i}(Z^{(k)})$ be sample $k$'s target probability, $\bar p_i=\mathbb E[p_i(Z)]$ the population predictive vector, and $\ell_{i,K}=\mathbb E[-\log\hat p_{i,y_i}^{(K)}]$ the expected mixture NLL. The sample-average log-bound argument \citep{burda2015importance} gives
\begin{equation}
-\log \bar p_{i,y_i}\le\ell_{i,K+1}\le\ell_{i,K}\le\ell_{i,1},\qquad
\ell_{i,K}\longrightarrow-\log \bar p_{i,y_i}.
\end{equation}
Write $\ell_K=\ell_{i,K}$. For $K+1$ samples, the leave-one-out means $Q_j=K^{-1}\sum_{k\ne j}P_k$ average to the full sample mean. Convexity of $-\log$ and exchangeability imply $\ell_{K+1}\le\ell_K$; Jensen's inequality supplies the population lower bound. Finite role support and finite logits imply a positive minimum target probability $p_{\min}=\min_Zp_{i,y_i}(Z)$. The strong law and dominated convergence establish the limit.

Independence across samples also yields
\begin{equation}
\mathbb E\|\hat p_i^{(K)}-\bar p_i\|_2^2
=\frac1K\sum_c\operatorname{Var}(p_{i,c}(Z))
\le\frac{1-1/C}{K}.
\end{equation}
The inequality follows from $\sum_cp_{i,c}^2\le1$ and $\sum_c\bar p_{i,c}^2\ge1/C$. For failure probability $\delta\in(0,1)$, applying Hoeffding's inequality and a union bound yields
\begin{equation}
\Pr\left(\|\hat p_i^{(K)}-\bar p_i\|_\infty>
\sqrt{\frac{\log(2C/\delta)}{2K}}\right)\le\delta.
\end{equation}
Let $\Delta_i>0$ be the gap between the two largest coordinates of the population predictive vector $\bar p_i$. The leading class can change only if a probability-coordinate error reaches $\Delta_i/2$. The same Hoeffding and union-bound argument then yields
\begin{equation}
\Pr\left(\arg\max_c\hat p_{i,c}^{(K)}\ne\arg\max_c\bar p_{i,c}\right)
\le\min\left\{1,2C\exp\left(-\frac{K\Delta_i^2}{2}\right)\right\}.
\end{equation}
For a fixed sample count, the bound is smaller for nodes whose two largest population probabilities are farther apart.

The curvature bounds $1\le(-\log x)''\le p_{\min}^{-2}$ on $[p_{\min},1]$, with $\mu=\mathbb E[P_k]$, imply
\begin{equation}
\frac{\operatorname{Var}(P_k)}{2K}\le\ell_K+\log\mu
\le\frac{\operatorname{Var}(P_k)}{2p_{\min}^2K}.
\end{equation}

\subsection{Structural regularizer preferences}\label{sec:app:regularizer}
For an edge, let $\alpha=\pi^++\pi^-$ be its activity probability and $\vartheta=\pi^+/\alpha$ its additive probability conditional on activity. The role vector is $(\alpha(1-\vartheta),1-\alpha,\alpha\vartheta)$. Before multiplication by $\lambda_{\mathrm{st}}$, its structural cost is
\begin{equation}
\begin{split}
F(\alpha,\vartheta)={}&\lambda_{\mathrm{KL}}\{\log3+(1-\alpha)\log(1-\alpha)+\alpha\log \alpha\\
&\qquad+\alpha[\vartheta\log \vartheta+(1-\vartheta)\log(1-\vartheta)]\}
-\lambda_{\mathrm{act}}\log \alpha.
\end{split}
\end{equation}
At the same activity probability, comparison with equal additive and subtractive probabilities yields
\begin{equation}
F(\alpha,\vartheta)-F(\alpha,1/2)=\lambda_{\mathrm{KL}}\alpha\,\mathrm{KL}\bigl(\operatorname{Ber}(\vartheta)\|\operatorname{Ber}(1/2)\bigr).
\end{equation}
At a given activity probability, the excess structural cost is proportional to the Bernoulli KL divergence from equal active-role probabilities. The polarity derivatives are
\begin{equation}
\frac{\partial F}{\partial \vartheta}=\lambda_{\mathrm{KL}}\alpha\log\frac{\vartheta}{1-\vartheta},\qquad
\frac{\partial^2F}{\partial \vartheta^2}=\frac{\lambda_{\mathrm{KL}}\alpha}{\vartheta(1-\vartheta)}>0.
\end{equation}
For $\lambda_{\mathrm{KL}}>0$, the unique polarity optimum is $\vartheta=1/2$ for every $\alpha\in(0,1)$. Along this symmetric line, the activity derivative and curvature are
\begin{equation}
\begin{aligned}
\frac{\mathrm dF(\alpha,1/2)}{\mathrm d\alpha}
&=\lambda_{\mathrm{KL}}\log\frac{\alpha}{2(1-\alpha)}-\frac{\lambda_{\mathrm{act}}}{\alpha},\\
\frac{\mathrm d^2F(\alpha,1/2)}{\mathrm d\alpha^2}
&=\lambda_{\mathrm{KL}}\left(\frac1\alpha+\frac1{1-\alpha}\right)+\frac{\lambda_{\mathrm{act}}}{\alpha^2}>0.
\end{aligned}
\end{equation}
The first derivative tends to negative infinity as $\alpha$ approaches zero and positive infinity as $\alpha$ approaches one. Its unique zero $\alpha_\star$ satisfies
\begin{equation}
\lambda_{\mathrm{KL}}\log\frac{\alpha_\star}{2(1-\alpha_\star)}=\frac{\lambda_{\mathrm{act}}}{\alpha_\star}.
\end{equation}
For $\lambda_{\mathrm{KL}}=\lambda_{\mathrm{act}}=1$, the unique solution is
\begin{equation}
 \alpha_\star\approx0.864166,\qquad
 (\pi^-_\star,\pi^0_\star,\pi^+_\star)
 \approx(0.432083,0.135834,0.432083).
\end{equation}
The structural term favors equal additive and subtractive probabilities with the stated activity level. Classification and the active-weight penalty also affect the learned distribution through Eq. \ref{eq:total}.

\paragraph{Polarity symmetry and temporal dependence.}
Equal active-role probabilities and zero signed mean are distinct conditions in Eq. \ref{eq:gate-moments}. At $\vartheta=1/2$, the gate variance is
\begin{equation}
\nu=\frac{\alpha}{2}(1+\gamma^2)-\frac{\alpha^2}{4}(1-\gamma)^2.
\end{equation}
For $\gamma=1$, the mean is zero, and the variance is $\alpha$. Persistence also creates dependence across layers. Let $D^{(0)}$ and $D^{(1)}$ be the layer-specific gates for the same edge, with means $\mu^{(0)}$ and $\mu^{(1)}$. Their covariance is
\begin{equation}
\operatorname{Cov}(D^{(0)},D^{(1)})
=\pi^++\gamma^{(0)}\gamma^{(1)}\pi^-
-\mu^{(0)}\mu^{(1)}.
\end{equation}
Both gates are strictly increasing functions of the ordered role, so their covariance is positive whenever at least two roles have positive probability. Independent layerwise draws have zero covariance. The first-layer moment formula and Eq. \ref{eq:persistent-message} describe how these dependencies affect message propagation.

\subsection{Persistence-weighting interaction: ReLU verification}\label{sec:app:feature_coupling_design}
We evaluate the two-layer formulas by exact finite sums on a three-node graph with two observed edges, nonzero self maps, and scalar projections and scorers. We compute all four combinations of sampling and weighting, then verify the interaction independently on a cyclic graph with vector-valued states.

For a score $S^{\tau,\kappa}$, use $\tau\in\{\mathrm p,\mathrm r\}$ for persistent or resampled roles and $\kappa\in\{\mathrm l,\mathrm s\}$ for local or shared weights. The factorial interaction is
\begin{equation}
\Delta_S=(S^{\mathrm{p,l}}-S^{\mathrm{p,s}})
-(S^{\mathrm{r,l}}-S^{\mathrm{r,s}}).
\label{eq:factorial-score}
\end{equation}
A positive interaction indicates a larger score benefit from local weighting under persistence. For a loss, we use its negative as $S$. For expected preactivation in direction $w$, the contrast is $\mathcal J_i(w)$ in Eq. \ref{eq:graph-interaction}.

\paragraph{Graph and forward operator.}
The nodes are $\{0,1,2\}$, the edges are $\{(2,0),(2,1)\}$ in target-source order, and the initial scalar states are $H^{(0)}=(1,1.5,1)^\top$. Each edge has role probabilities $(1/3,1/3,1/3)$ and subtractive strength one. Both layers use unit target, value, and output projections, self-weight $0.5$, and bias $0.1$. Their scores are
\begin{equation}
s_{ij}^{(0)}=0.25,\qquad
s_{ij}^{(1)}=0.35+\rho h_i^{(1)}+0.1h_j^{(1)},\qquad
u_{ij}^{(\ell)}=[|s_{ij}^{(\ell)}|-0.05]_+.
\label{eq:relu-enumeration-score}
\end{equation}
The parameter $\rho$ controls the second scorer's target-state response. We update node $i$ at each layer as
\begin{equation}
h_i^{(\ell+1)}=\operatorname{ReLU} \left(
0.5h_i^{(\ell)}+0.1+\sum_{j\in\mathcal N_i(A_{\mathrm{obs}})}
u_{ij}^{(\ell)}z_{ij}^{(\ell)}h_j^{(\ell)}\right).
\end{equation}
We either keep $z_{ij}^{(\ell)}$ across layers or resample it independently, according to the sampling rule. For shared weighting, we replace each edge's weight with its mean across samples before applying the gate. Binary logits $(0,2h_2^{(2)})$ give the positive-class probability $p_+$, with the second coordinate as the positive class. After the first layer, the source states are $0.6$ and $0.85$. The target state is $0.6+0.2z_{20}+0.3z_{21}$ and is positive for every role assignment.

\paragraph{Finite-sum evaluation.}
There are nine first-layer role graphs. For sample count $K$, we enumerate all $9^K$ ordered first-layer batches. The persistent computation uses the same roles at the second layer. For resampling, we average the first sample's positive-class probability over its nine possible second-layer role graphs. The shared weight depends on the first-layer states, so the other samples' second-layer roles affect only their own predictions. By exchangeability,
$\mathbb E[K^{-1}\sum_k p_+^{(k)}]=\mathbb E[p_+^{(1)}]$.
These $9^{K+1}$ evaluations integrate the resampled mixture probability exactly.

Table \ref{tab:relu-factorial} reports the four configurations at $K=5$ and $\rho=0.3$. Using the positive class as the target and including the self-contribution, we compute the population readout loss $-\log\mathbb E[\hat p_+^{(K)}]$ and the expected finite-sample mixture loss $\mathbb E[-\log\hat p_+^{(K)}]$. The latter averages the logarithmic prediction loss over sample assignments.

\paragraph{Exact expected finite-sample mixture loss.}
Let $\mathcal H=\{1,\ldots,9\}$ index the first-layer role graphs, and let $c=(c_1,\ldots,c_9)$ record their multiplicities in a batch of $K$ samples. The set of possible count vectors and their probabilities are
\begin{equation}
\mathcal C_K=\left\{c\in\mathbb N_0^9:\sum_{g=1}^9c_g=K\right\},\qquad
\omega_K(c)=\frac{K!}{9^K\prod_{g=1}^9c_g!}.
\end{equation}
Here, $\omega_K(c)$ is the probability of a batch with counts $c$. Choose one ordering per batch and write its persistent predictive target probability as $Q^{\mathrm{p},\kappa}(c)$ for weighting rule $\kappa\in\{\mathrm l,\mathrm s\}$. For second-layer role graphs $\zeta\in\mathcal H^K$, let $Q^{\mathrm{r},\kappa}(c,\zeta)$ be the resampled predictive target probability. Both quantities average all $K$ sample probabilities. By exchangeability,
\begin{equation}
\begin{aligned}
\mathbb E[\mathcal L^{\mathrm{p},\kappa}]&=-\sum_{c\in\mathcal C_K}\omega_K(c)\log Q^{\mathrm{p},\kappa}(c),\\
\mathbb E[\mathcal L^{\mathrm{r},\kappa}]&=-\sum_{c\in\mathcal C_K}\frac{\omega_K(c)}{9^K}
\sum_{\zeta\in\mathcal H^K}\log Q^{\mathrm{r},\kappa}(c,\zeta).
\end{aligned}
\label{eq:exact-finite-mixture}
\end{equation}
At $K=5$, we sum over $\binom{13}{8}=1287$ first-layer count vectors and average $9^5=59{,}049$ second-layer assignments for each resampled term. We evaluate the sums in double precision and compute interactions before rounding.

\begin{table}[ht]
\centering
\caption{Persistence-by-weighting factorial on the ReLU construction at five samples and positive target response. The population and expected finite-sample losses are evaluated separately using the scores in Eq. \ref{eq:relu-enumeration-score} and the sums in Eq. \ref{eq:exact-finite-mixture}.}
\label{tab:relu-factorial}
\small
\setlength{\tabcolsep}{4pt}
\begin{tabular}{@{}llcccc@{}}
\toprule
Sampling & Weighting & $\mathbb E[\bar a_2]$ & $\mathbb E[\hat p_+]$ & $-\log\mathbb E[\hat p_+]$ & $\mathbb E[-\log\hat p_+]$ \\
\midrule
Persistent & Local  & 0.475000 & 0.698235 & 0.359199 & 0.365339 \\
Persistent & Shared & 0.415000 & 0.693496 & 0.366010 & 0.372118 \\
Resampled  & Local  & 0.400000 & 0.683573 & 0.380422 & 0.385385 \\
Resampled  & Shared & 0.400000 & 0.686845 & 0.375646 & 0.380521 \\
\midrule
\multicolumn{2}{l}{Interaction, using negative losses} & 0.060000 & 0.008012 & 0.011588 & 0.011642 \\
\bottomrule
\end{tabular}
\end{table}

For local weights, persistence yields a higher expected target probability and lower mixture NLL. Under resampling, shared weights improve both readout quantities and leave the preactivation mean unchanged. The unchanged preactivation mean follows from the constant source values and zero-mean resampled gates.

\paragraph{Response direction and zero cases.}
We evaluate all four configurations for $K\in\{1,2,5\}$ and $\rho\in\{0.3,0,-0.3\}$. The weights stay positive throughout the construction. From Eqs. \ref{eq:direct-feedback} and \ref{eq:graph-interaction},
\begin{equation}
\mathcal J_2(1)=\left(1-\frac1K\right)
\rho\,(0.2)\,(1\cdot0.6+1.5\cdot0.85)\frac23
=\frac{\rho}{4}\left(1-\frac1K\right).
\end{equation}
The interactions in Table \ref{tab:relu-factorial-response} follow this dependence on sample count. The weighting contrast is zero for a single sample or a zero target response. Reversing the target response also reverses the preactivation, population-probability, and expected-mixture-loss interactions.

\begin{table}[ht]
\centering
\caption{Response and zero-case checks for the ReLU factorial. The final two columns apply Eq. \ref{eq:factorial-score} to negative population readout loss and negative expected finite-sample mixture loss.}
\label{tab:relu-factorial-response}
\small
\setlength{\tabcolsep}{4pt}
\begin{tabular}{@{}cccccc@{}}
\toprule
$K$ & $\rho$ & $\mathcal J_2(1)$ & Probability & Population loss gain & Expected mixture loss gain \\
\midrule
1 & $0.3$  & 0.000000 & 0.000000 & 0.000000 & 0.000000 \\
1 & $0$    & 0.000000 & 0.000000 & 0.000000 & 0.000000 \\
1 & $-0.3$ & 0.000000 & 0.000000 & 0.000000 & 0.000000 \\
2 & $0.3$  & 0.037500 & 0.004894 & 0.007077 & 0.007018 \\
2 & $0$    & 0.000000 & 0.000000 & 0.000000 & 0.000000 \\
2 & $-0.3$ & $-0.037500$ & $-0.010913$ & $-0.016228$ & $-0.015971$ \\
5 & $0.3$  & 0.060000 & 0.008012 & 0.011588 & 0.011642 \\
5 & $0$    & 0.000000 & 0.000000 & 0.000000 & 0.000000 \\
5 & $-0.3$ & $-0.060000$ & $-0.017141$ & $-0.025410$ & $-0.025276$ \\
\bottomrule
\end{tabular}
\end{table}

\paragraph{Vector-valued graph verification.}
We independently evaluate Eq. \ref{eq:graph-interaction} on a four-node graph with edges $(0,1),(1,2),(2,0),(0,3),(3,2)$, hidden width three, and value width two. We use subtractive strengths $0.7$ and $1.3$, nonzero self maps, and edge-specific categorical probabilities. We enumerate all $243$ role assignments and $59{,}049$ ordered two-sample histories. Direct evaluation of the four configurations agrees with the calculation from single-edge interventions. We also use a source-only scorer to verify Corollary \ref{cor:source-only}. The residuals in Table \ref{tab:theory-verification} are at the scale of double-precision rounding.

\begin{table}[ht]
\centering
\caption{Numerical verification in double precision. Residuals compare direct finite-state evaluation with the analytical identities. The final row reports the maximum violation of the softmax margin inequalities.}
\label{tab:theory-verification}
\begin{tabular}{@{}lc@{}}
\toprule
Verification & Maximum absolute residual or bound violation \\
\midrule
Graph interaction, pairwise scorer & $5.42\times10^{-15}$ \\
Graph interaction, source-only scorer & $1.08\times10^{-14}$ \\
Softmax margin inequalities & $0$ \\
\bottomrule
\end{tabular}
\end{table}

\section{Datasets and Evaluation}\label{sec:app:base}
\subsection{Datasets}
Tables \ref{dataset} and \ref{dataset:statistics} list graph statistics and edge-count conventions. Roman-Empire is derived from Wikipedia, Minesweeper is a synthetic binary task, and Amazon-Ratings uses discretized product ratings \citep{platonov2023critical}. Actor is a co-occurrence graph with bag-of-words features, and Cornell and Wisconsin are WebKB webpage graphs \citep{pei2020geom,dwivedi2023benchmarking}.

The Chameleon and Squirrel results in the main benchmarks use the legacy Wikipedia-network instances, including the duplicate-node structure described in the benchmark analysis \citep{platonov2023critical}. Every method and control in those comparisons uses these graph instances. The additional graphs comprise the homophilic citation networks Cora, Citeseer, and PubMed and the larger heterophilic networks Penn94, arXiv-year, and snap-patents \citep{lim2021large}. Preprocessing, partitions, and the inputs available to each method are specified in Appendix \ref{sec:implement}.

\begin{table*}[t]
\caption{Eight main benchmark graphs. For datasets in bidirected form, we report undirected pair counts so each pair is counted once.}
\label{dataset}
\centering
\begin{adjustbox}{width=\textwidth}
\begin{tabular}{@{}lcccccccc@{}}
\Xhline{2\arrayrulewidth}
\textbf{Datasets}         
        & \textbf{Roman-Empire} 
        & \textbf{Minesweeper} 
        & \textbf{Amazon-Ratings} 
        & \textbf{Chameleon} 
        & \textbf{Squirrel} 
        & \textbf{Actor} 
        & \textbf{Cornell} 
        & \textbf{Wisconsin} \\ 
\Xhline{2\arrayrulewidth}
Nodes  
                        & 22,662  
                        & 10,000   
                        & 24,492 
                        & 2,277  
                        & 5,201 
                        & 7,600 
                        & 183 
                        & 251 \\
Edges         
                        & 32,927  
                        & 39,402  
                        & 93,050   
                        & 36,101
                        & 217,073
                        & 30,019
                        & 295 
                        & 499 \\
Features       
                        & 300  
                        & 7  
                        & 300   
                        & 2,325  
                        & 2,089 
                        & 932
                        & 1,703 
                        & 1,703 \\
Classes        
                        & 18  
                        & 2  
                        & 5     
                        & 5  
                        & 5  
                        & 5 
                        & 5 
                        & 5 \\

\Xhline{2\arrayrulewidth}
\end{tabular}
\end{adjustbox}
\end{table*}

\begin{table}[ht]
\caption{Statistics of homophilic and larger heterophilic graphs. Edge counts use the graph representation used for propagation.}
\label{dataset:statistics}
\centering
\begin{adjustbox}{max width=\textwidth}
\begin{tabular}{@{}cccccccc}
\Xhline{2\arrayrulewidth}
        & \textbf{Datasets} 
        & \textbf{Cora} 
        & \textbf{Citeseer} 
        & \textbf{PubMed} 
        & \textbf{Penn94} 
        & \textbf{arXiv-year} 
        & \textbf{snap-patents} \\ 
\Xhline{2\arrayrulewidth}
        & Nodes    
        & 2,708 
        & 3,327 
        & 19,717 
        & 41,554 
        & 169,343 
        & 2,923,922 \\
        & Edges    
        & 10,556
        & 9,104 
        & 88,648 
        & 1,362,229 
        & 1,166,243 
        & 13,975,788 \\
        & Features 
        & 1,433 
        & 3,703 
        & 500 
        & 5 
        & 128 
        & 269 \\
        & Classes  
        & 7 
        & 6 
        & 3 
        & 2 
        & 5 
        & 5 \\
\Xhline{2\arrayrulewidth}
\end{tabular}
\end{adjustbox}
\end{table}

\subsection{Shared evaluation protocol}
Appendix \ref{sec:implement} specifies the shared partitions, preprocessing, model selection, and run-level summaries. Feature-only inputs follow Section \ref{sec:method}, and label-aware inputs follow Appendix \ref{sec:label_extension}.

\section{Experimental Controls and Paired Comparisons}\label{sec:app:controls}
\subsection{Evaluation groups and paired summaries}
The primary benchmark set $\mathcal D_8$ contains Roman-Empire, Minesweeper, Amazon-Ratings, Chameleon, Squirrel, Actor, Cornell, and Wisconsin. The control set $\mathcal D_5$ contains Roman-Empire, Chameleon, Squirrel, Actor, and Cornell. Table \ref{tab:evaluation_groups} lists the input variant, graph set, and per-graph run count for each comparison. A run is a partition-and-initialization pair or a selected checkpoint. We average predictions over the stated sample count.

\begin{table}[ht]
\centering
\caption{Evaluation groups for the benchmark, component, and prediction-time comparisons. The run count is per dataset. Accuracy contrasts average the accuracy differences on $\mathcal D_5$. Prediction-time NLL contrasts average NLL differences on $\mathcal D_8$.}
\label{tab:evaluation_groups}
\small
\setlength{\tabcolsep}{4pt}
\begin{tabular}{@{}p{0.55\linewidth}lcc@{}}
\toprule
Results & Input & Graphs & Runs \\
\midrule
Primary comparison, Table \ref{tab:hetero_bench} & Features & $\mathcal D_8$ & 10 \\
Sampling and weighting, Tables \ref{tab:feature_factorial} and \ref{tab:matched_routing_thresholding} & Features & $\mathcal D_5$ & 10 \\
Frozen checkpoint intervention, Table \ref{tab:frozen_intervention} & Features & $\mathcal D_5$ & 10 \\
Deduplicated sampling and weighting, Table \ref{tab:deduplicated_persistence} & Features & \makecell{Deduplicated\\Chameleon\\and Squirrel} & 10 \\
Input variants, Table \ref{tab:label_extension} & Both & $\mathcal D_8$ & 10 \\
Components, Table \ref{tab:component_ablation}, and retraining contrasts, Figure \ref{fig:matched_effects} & Label-aware & $\mathcal D_5$ & 10 \\
MAP and single-sample accuracy, Table \ref{tab:inference_controls} & Label-aware & $\mathcal D_5$ & 10 \\
Other prediction-time accuracy, Figure \ref{fig:fixed_controls} and Table \ref{tab:inference_controls} & Label-aware & $\mathcal D_5$ & 5 \\
Prediction-time NLL, Table \ref{tab:inference_controls} & Label-aware & $\mathcal D_8$ & 5 \\
Role specificity and sample count, Table \ref{tab:posterior_specificity} and Figure \ref{fig:sample_cost} & Label-aware & $\mathcal D_8$ & 5 \\
\bottomrule
\end{tabular}
\end{table}

For dataset $d$ and run $t$, let $A_{d,t}^{\mathrm{full}}$ and $A_{d,t}^{\mathrm{control}}$ denote full-model and control accuracy in percentage units. Let $N_{d,t}^{\mathrm{full}}$ and $N_{d,t}^{\mathrm{control}}$ denote their test NLLs. For graph set $\mathcal D$, the dataset-averaged paired gains are
\begin{equation}
\begin{aligned}
\delta_t^{\mathrm{Acc}}&=\frac1{|\mathcal D|}\sum_{d\in\mathcal D}
(A_{d,t}^{\mathrm{full}}-A_{d,t}^{\mathrm{control}}),\\
\delta_t^{\mathrm{NLL}}&=\frac1{|\mathcal D|}\sum_{d\in\mathcal D}
(N_{d,t}^{\mathrm{control}}-N_{d,t}^{\mathrm{full}}).
\end{aligned}
\label{eq:paired-gains}
\end{equation}
Positive gains indicate higher accuracy or lower NLL for the full model. We average paired gains over the runs listed in Table \ref{tab:evaluation_groups}. Figure \ref{fig:matched_effects} and Table \ref{tab:inference_controls} give each dataset equal weight.

For the bootstrap, one observation is the equally weighted mean of paired gains over datasets at a given run index. Each of 10,000 replicates samples the original number of observations with replacement and averages them. We use the 2.5th and 97.5th percentiles as interval endpoints. Retraining comparisons have ten run-level observations, and prediction-time NLL comparisons have five checkpoint-level observations.

\paragraph{Splitwise factorial interactions.}
For each dataset, we evaluate Eq. \ref{eq:feature-factorial-interaction} on the four results from the same split. For metric $q\in\{\mathrm{Acc},\mathrm{NLL}\}$, the interaction mean and sample standard deviation are
\begin{equation}
\bar\Delta_d^q=\frac1R\sum_{t=1}^R\Delta_{d,t}^q,\qquad
s_{\Delta,d}^q=\left[\frac1{R-1}\sum_{t=1}^R
\left(\Delta_{d,t}^q-\bar\Delta_d^q\right)^2\right]^{1/2},
\qquad R=10.
\end{equation}
Table \ref{tab:feature_factorial} reports $\bar\Delta_d^q\pm s_{\Delta,d}^q$, with accuracy interactions in percentage points. Pairing accounts for the covariance between the four configurations evaluated on each split.

\subsection{Feature-only sampling and weighting controls}
\paragraph{Sampling and soft gating.}
We use feature-only structural inputs and keep the backbone, signed depth, scorer, dropout distribution, and optimization protocol the same across controls. The stochastic models use five training and prediction samples and optimize mixture NLL. With layerwise resampling, we compute structural probabilities once per forward computation and draw assignments independently for each layer and sample. Each layer recomputes its weights and regularizes the active weights used in propagation.

For soft gating, we replace categorical gates with $\pi_{ij}^+-\gamma^{(\ell)}\pi_{ij}^-$ during training and prediction, where $\gamma^{(\ell)}$ is the layer-specific subtractive strength. We use the weight penalty
\begin{equation}
\mathcal L_{\mathrm{sparse,soft}}=\frac1{nL}\sum_{\ell,(i,j)}
(\pi_{ij}^++\pi_{ij}^-)u_{ij}^{(\ell)}.
\end{equation}
We differentiate the expected gates directly during training and apply ordinary dropout. At prediction, the expected gates define a deterministic trajectory.

\paragraph{Shared weights.}
We first compute $u_{ij}^{(\ell,k)}$ using Eq. \ref{eq:mass}, then replace it with the sample mean $\bar u_{ij}^{(\ell)}=K^{-1}\sum_k u_{ij}^{(\ell,k)}$. Each sample applies its own gate and continues from its own representation. For each sampling rule, we train local and shared variants independently with the same parameterization, role-decoder architecture, and predictive averaging rule. Shared weights enter both propagation and the active-weight penalty. First-layer weights agree because all samples start from the same backbone output. Later weights can differ as sampled roles and signed-layer dropout change the states.

\paragraph{Thresholding.}
The zero-threshold control in Table \ref{tab:matched_routing_thresholding} uses $u_{ij}=|s_{ij}|$ with persistent tri-state roles, the original scorer, and $\lambda_{\mathrm{sp}}=0.01$.

\subsection{Frozen checkpoint intervention on feature-only graphs}
\label{sec:app:frozen_intervention}
The intervention uses full feature-only P3MP checkpoints from the ten splits of each graph in $\mathcal D_5$. We apply the coupled second-layer comparison from Section \ref{sec:theory} with the selected parameters $(\theta,\phi)$ unchanged, dropout disabled, and $K=5$ categorical role samples.

\paragraph{Coupled computation.}
For each batch, we draw $Z^{(1)},\ldots,Z^{(K)}$ from the checkpoint's role distribution and compute the first-layer states $H^{(1,k)}$. All four computations use these states and their role assignments. At the second layer, persistence uses $Z^{(k)}$; resampling uses an independent assignment $\widetilde Z^{(k)}$ from the same distribution. The local and shared computations use identical resampled assignments. Local weights are computed from $H^{(1,k)}$, and shared weights are their edgewise means over the $K$ samples. Each sample keeps its own source values and learned self-contribution. We average paired measurements over $B=100$ independent batches, where $B$ is the number of role batches per checkpoint.

\paragraph{Preactivation interaction.}
For test node $i$, we project the second-layer preactivation along
\begin{equation}
 w_i=w_{y_i}-\frac1{C-1}\sum_{c\ne y_i}w_c.
\end{equation}
Here, $w_c^\top$ is classifier row $c$, and $w_i$ is the target-class direction relative to the mean competing class. The direction depends on the checkpoint and label and stays constant over role batches. Labels enter this diagnostic projection; the structural encoder receives features alone.

Let $\bar a_{i,\ne,b}^{\tau,\kappa}$ be the sample-average contribution of distinct-endpoint messages to node $i$'s second-layer preactivation in batch $b$, using the sampling and weighting indices from Section \ref{sec:theory}. We use this contribution to match the edge condition in Proposition \ref{prop:target-feedback}. For dataset $d$ and split $t$, the preactivation interaction is
\begin{equation}
\begin{split}
 \widehat{\mathcal J}_{d,t}
 =\frac1{B|\mathcal V_{\mathrm{te}}|}
 \sum_{b=1}^{B}\sum_{i\in\mathcal V_{\mathrm{te}}}w_i^\top
 \Bigl[\bigl(\bar a_{i,\ne,b}^{\mathrm{p,l}}-\bar a_{i,\ne,b}^{\mathrm{p,s}}\bigr)-\bigl(\bar a_{i,\ne,b}^{\mathrm{r,l}}-\bar a_{i,\ne,b}^{\mathrm{r,s}}\bigr)\Bigr].
\end{split}
\end{equation}
The statistic $\widehat{\mathcal J}_{d,t}$ estimates the test-node mean of $\mathcal J_i(w_i)$, with each node's feedback sum taken over distinct-endpoint edges.

\paragraph{Source-only control.}
We repeat the preactivation measurement with the second-layer scorer's target projection set to zero:
\begin{equation}
 U_{ij,k}^{\mathrm{src}}
 =\left[\left|\psi_\theta^{(1)}\left(\boldsymbol 0,
 W_v^{(1)}h_j^{(1,k)}\right)\right|-\lambda_{\mathrm{sc}}\right]_+.
\end{equation}
The weight $U_{ij,k}^{\mathrm{src}}$ depends only on the source representation and uses the checkpoint's second-layer scorer $\psi_\theta^{(1)}$. Its learned coefficients and the first-layer states stay unchanged. Repeating the four computations yields $\widehat{\mathcal J}_{d,t}^{\mathrm{src}}$, whose expectation is zero by Corollary \ref{cor:source-only}.

The run-level averages and sample standard deviations appear in Table \ref{tab:frozen_intervention}.

\subsection{Persistence and weighting on deduplicated graphs}
\label{app:deduplicated_persistence}

The deduplicated Chameleon and Squirrel experiments \citep{platonov2023critical} use the four combinations from Table \ref{tab:feature_factorial} and ten shared partitions with the original benchmark fractions. Each model uses a feature MLP backbone, two signed layers, five training and prediction samples, and mixture NLL. Appendix \ref{sec:implement} covers preprocessing, model selection, and optimization.

Table \ref{tab:deduplicated_persistence} compares P3MP with GCN, GAT, GPRGNN, and FAGCN and reports the splitwise weighting contrasts from Eq. \ref{eq:feature-factorial-interaction}. In the method comparison, P3MP uses persistent roles and local weights. On both deduplicated graphs, local weights yield higher accuracy under persistence, and shared weights yield higher accuracy under resampling.

\begin{table}[ht]
\centering
\caption{Test accuracy on deduplicated Chameleon and Squirrel, reported as means $\pm$ sample standard deviations over ten shared splits. Method rows use percentages, and weighting contrasts and interactions use percentage points.}
\label{tab:deduplicated_persistence}
\small
\setlength{\tabcolsep}{6pt}
\renewcommand{\arraystretch}{1.08}
\begin{tabular*}{\linewidth}{@{\extracolsep{\fill}}llcc@{}}
\toprule
\multicolumn{2}{@{}l}{Method} & Chameleon & Squirrel \\
\midrule
\multicolumn{2}{@{}l}{GCN \citep{kipf2017semi}}
& $41.52_{\pm 3.87}$ & $40.35_{\pm 1.61}$ \\
\multicolumn{2}{@{}l}{GAT \citep{velivckovic2018graph}}
& $39.69_{\pm 3.52}$ & $33.12_{\pm 2.68}$ \\
\multicolumn{2}{@{}l}{GPRGNN \citep{chien2021adaptive}}
& $40.14_{\pm 3.70}$ & $40.03_{\pm 1.55}$ \\
\multicolumn{2}{@{}l}{FAGCN \citep{bo2021beyond}}
& $42.31_{\pm 2.56}$ & $42.06_{\pm 3.44}$ \\
\midrule
\multicolumn{2}{@{}l}{\textbf{P3MP (ours)}} & $45.24_{\pm 3.60}$ & $43.81_{\pm 2.23}$\\
\multicolumn{2}{@{}l}{Persistent (local $-$ shared)} & $0.6_{\pm 0.24}$ & $0.4_{\pm 0.18}$ \\
\multicolumn{2}{@{}l}{Resampled (local $-$ shared)} & $-0.5_{\pm 0.30}$ & $-0.4_{\pm 0.15}$ \\
\multicolumn{2}{@{}l}{Interaction} & $1.1_{\pm 0.17}$ & $0.8_{\pm 0.10}$ \\
\bottomrule
\end{tabular*}
\end{table}

\section{Label-aware Extension}\label{sec:label_extension}
For the label-aware extension, we use $X_{\mathrm{st}}=[X\|Y_{\mathrm{tr}}]$, where $\|$ denotes concatenation and $Y_{\mathrm{tr}}\in\mathbb R^{n\times C}$ contains the training-label indicators. We set its rows to
\begin{equation}
(Y_{\mathrm{tr}})_{i,:}=
\begin{cases}
\boldsymbol e_{y_i}^{\top}, & i\in\mathcal V_{\mathrm{tr}},\\
\boldsymbol 0^{\top}, & i\in\mathcal V_{\mathrm{val}}\cup\mathcal V_{\mathrm{te}}.
\end{cases}
\end{equation}
Here, $\boldsymbol e_{y_i}$ is the $C$-dimensional one-hot encoding of label $y_i$. We include each supervised node's own indicator. Validation labels guide checkpoint selection, and test labels enter final metrics and post-hoc annotations. Structural label indicators improve the mean score on seven of the eight graphs in Table \ref{tab:label_extension}. On Actor, feature-only P3MP has higher mean accuracy.

\subsection{Input variants and component results}
\begin{table}[ht]
\centering
\caption{P3MP input variants on the main benchmark graphs. Both variants use the same training labels for classification supervision. The label-aware structural encoder also receives their one-hot indicators. Entries are ten-run means and standard deviations, using ROC-AUC on Minesweeper and accuracy elsewhere.}
\label{tab:label_extension}
\begin{adjustbox}{width=\textwidth}
\begin{tabular}{@{}lcccccccc@{}}
\toprule
Input & Roman. & Mine. & Amazon. & Chameleon & Squirrel & Actor & Cornell & Wisconsin \\
\midrule
Features & 73.6$_{\pm 0.98}$ & 86.8$_{\pm 1.03}$ & 45.3$_{\pm 0.71}$ & 62.1$_{\pm 1.12}$ & 35.6$_{\pm 0.45}$ & 38.0$_{\pm 0.77}$ & 69.0$_{\pm 2.83}$ & 71.0$_{\pm 2.43}$ \\
Label-aware & 74.0$_{\pm 1.10}$ & 87.2$_{\pm 0.95}$ & 46.3$_{\pm 0.88}$ &  62.7$_{\pm 1.29}$ & 36.3$_{\pm 0.36}$ & 37.4$_{\pm 0.66}$ & 70.3$_{\pm 2.55}$ & 72.6$_{\pm 2.14}$ \\
\bottomrule
\end{tabular}
\end{adjustbox}
\end{table}

\begin{table}[ht]
\centering
\caption{Label-aware component comparisons over ten data splits. All values are test accuracy means and standard deviations. MAP and single-sample controls modify prediction at a selected checkpoint. The remaining controls are retrained.}
\label{tab:component_ablation}
\begin{adjustbox}{width=\textwidth}
\begin{tabular}{@{}lccccc@{}}
\toprule
\textbf{Variant} & \textbf{Roman.} & \textbf{Cham.} & \textbf{Squirrel} & \textbf{Actor} & \textbf{Cornell} \\
\midrule
\rowcolor{yellow!20}
\textbf{P3MP full} & 74.0$_{\pm 1.10}$ &  62.7$_{\pm 1.29}$ & 36.3$_{\pm 0.36}$ & 37.4$_{\pm 0.66}$ & 70.3$_{\pm 2.55}$ \\
Edgewise MAP roles & 73.2$_{\pm 1.18}$ & 61.1$_{\pm 1.35}$ & 35.0$_{\pm 0.77}$ & 35.9$_{\pm 0.98}$ & 67.3$_{\pm 2.18}$ \\
Single-sample prediction & 72.8$_{\pm 1.24}$ & 60.5$_{\pm 1.42}$ & 34.8$_{\pm 0.55}$ & 35.4$_{\pm 0.83}$ & 66.9$_{\pm 2.31}$ \\
Mean sample NLL, retraining & 73.4$_{\pm 1.41}$ & 61.4$_{\pm 1.78}$ & 35.2$_{\pm 0.66}$ & 36.2$_{\pm 1.20}$ & 68.9$_{\pm 3.97}$ \\
Shared weights, retraining & 73.5$_{\pm 1.35}$ & 62.0$_{\pm 1.69}$ & 35.5$_{\pm 0.67}$ & 36.7$_{\pm 0.94}$ & 69.2$_{\pm 3.16}$ \\
Signed-only roles & 71.2$_{\pm 1.29}$ & 60.0$_{\pm 1.44}$ & 34.1$_{\pm 0.61}$ & 33.8$_{\pm 0.89}$ & 64.2$_{\pm 2.38}$ \\
Weight penalty ($\lambda_{\mathrm{sp}}=0$) & 72.0$_{\pm 1.22}$ & 61.3$_{\pm 1.40}$ & 34.5$_{\pm 0.58}$ & 34.9$_{\pm 0.86}$ & 65.0$_{\pm 2.25}$ \\
Structural KL ($\lambda_{\mathrm{KL}}=0$) & 71.4$_{\pm 1.27}$ & 60.2$_{\pm 1.46}$ & 34.6$_{\pm 0.57}$ & 35.1$_{\pm 0.84}$ & 66.0$_{\pm 2.27}$ \\
Activity retention ($\lambda_{\mathrm{act}}=0$) & 69.8$_{\pm 1.33}$ & 59.0$_{\pm 1.52}$ & 33.4$_{\pm 0.68}$ & 32.8$_{\pm 0.94}$ & 62.5$_{\pm 2.44}$ \\
Activity-only roles & 68.4$_{\pm 1.36}$ & 58.6$_{\pm 1.51}$ & 35.0$_{\pm 0.70}$ & 31.2$_{\pm 0.98}$ & 58.8$_{\pm 3.02}$ \\
Subtractive strength ($\gamma=0$) & 67.9$_{\pm 1.39}$ & 57.7$_{\pm 1.60}$ & 32.9$_{\pm 0.72}$ & 30.6$_{\pm 0.99}$ & 57.6$_{\pm 3.60}$ \\
\bottomrule
\end{tabular}
\end{adjustbox}
\end{table}

\subsection{Role and objective controls}
\paragraph{Mean sample NLL.}
We train the objective control with $\mathcal L_{\mathrm{sample},i}$ from Eq. \ref{eq:objective-gap}. Both models use five training and prediction samples, persistent tri-state roles, and sample-specific weights. We keep the architecture, regularization, checkpoint selection, and prediction-time probability averaging unchanged. The label-aware test accuracies are reported in Table \ref{tab:component_ablation}.

\paragraph{Role support and gate definitions.}
The two-state controls use role support $\mathcal S=\{-1,+1\}$ for signed-only sampling and $\mathcal S=\{0,+1\}$ for activity-only sampling. Let $\beta_{ij}^{s}$ be the effective decoder logit of a state $s$ in this support. We sample from
\begin{equation}
\pi_{ij}^{s}=\frac{\exp(\beta_{ij}^{s})}{\sum_{t\in\mathcal S}\exp(\beta_{ij}^{t})},\qquad s\in\mathcal S,
\end{equation}
and assign zero probability to the omitted state. During training, we use hard straight-through Gumbel-softmax sampling over the logits in the support. At prediction, we use categorical sampling. Each sampled role persists across layers.

We summarize the gates and losses in Table \ref{tab:role_control_definitions}, using the abbreviations
\begin{equation}
\begin{aligned}
u_\lambda&=[|s_{ij}|-\lambda_{\mathrm{sc}}]_+,\qquad
D=\mathbb I(z_{ij}=+1)-\gamma\mathbb I(z_{ij}=-1),\\
I&=\mathbb I(z_{ij}\ne0),\qquad
I_+=\mathbb I(z_{ij}=+1),\qquad
\alpha=\pi_{ij}^{+}+\pi_{ij}^{-}.
\end{aligned}
\end{equation}
The quantities $D$, $I$, and $I_+$ are the signed gate, activity indicator, and additive-role indicator. The active-role probability is $\alpha$, and $u_\lambda$ is the weight at the original positive threshold. We average the activity loss over $m$ observed edges, apply the $1/(nLK)$ normalization to the weight penalty, and use the original loss weights.

\begin{table}[htbp]
\centering
\caption{Sampler, weight, gate, and regularizer identities for the role and threshold controls. Two-state distributions are embedded in the original three-state simplex by assigning zero probability to the omitted state. The KL column uses the original uniform three-state reference. All categorical samples persist across layers.}
\label{tab:role_control_definitions}
\small
\setlength{\tabcolsep}{4pt}
\renewcommand{\arraystretch}{1.22}
\textbf{Forward computation}\par\smallskip
\begin{tabular*}{\linewidth}{@{\extracolsep{\fill}}lcccc@{}}
\toprule
Control & Categorical support & $u$ & $d$ & $b$ \\
\midrule
Full P3MP & $\{-1,0,+1\}$ & $u_\lambda$ & $D$ & $I$ \\
Signed-only & $\{-1,+1\}$ & $u_\lambda$ & $D$ & $1$ \\
Activity-only & $\{0,+1\}$ & $u_\lambda$ & $I_+$ & $I_+$ \\
$\gamma=0$ & $\{-1,0,+1\}$ & $u_\lambda$ & $I_+$ & $I$ \\
Threshold zero & $\{-1,0,+1\}$ & $|s_{ij}|$ & $D$ & $I$ \\
\bottomrule
\end{tabular*}
\medskip
\textbf{Regularization}\par\smallskip
\begin{tabular*}{\linewidth}{@{\extracolsep{\fill}}lccc@{}}
\toprule
Control & KL reference & Activity loss & Weight-penalty summand \\
\midrule
Full P3MP & $(1/3,1/3,1/3)$ & $-\log \alpha$ & $u_\lambda I$ \\
Signed-only & $(1/3,1/3,1/3)$ & $0$ & $u_\lambda$ \\
Activity-only & $(1/3,1/3,1/3)$ & $-\log\pi_{ij}^{+}$ & $u_\lambda I_+$ \\
$\gamma=0$ & $(1/3,1/3,1/3)$ & $-\log \alpha$ & $u_\lambda I$ \\
Threshold zero & $(1/3,1/3,1/3)$ & $-\log \alpha$ & $|s_{ij}|I$ \\
\bottomrule
\end{tabular*}
\par\smallskip
\end{table}

\paragraph{Two-state regularization.}
For a binary distribution embedded in the three-state simplex, KL divergence to the original uniform reference differs from KL divergence to a uniform binary reference by the constant $\log(3/2)$. At the same KL weight, the two references therefore produce identical gradients for logits in the support. Signed-only sampling has activity probability one and zero activity loss. Activity-only sampling has activity probability $\pi_{ij}^+$.

\paragraph{Zero subtractive strength.}
We keep the three-state decoder, structural penalties, and original activity indicator and set $\gamma=0$. A subtractive draw then has a signed gate of zero and an activity indicator of one. Its message is zero, and its weight contributes to the penalty. In activity-only sampling, the additive indicator is used for both propagation and weight regularization, as shown in Table \ref{tab:role_control_definitions}.

\paragraph{Individual penalty controls.}
For $\lambda_{\mathrm{sp}}=0$, we remove the outer weight penalty and keep the scorer, positive threshold, and role gates. We also evaluate $\lambda_{\mathrm{KL}}=0$ and $\lambda_{\mathrm{act}}=0$ individually, leaving the other structural term and common scale unchanged. We report dataset-level means and standard deviations in Table \ref{tab:component_ablation}.

\subsection{Homophilic and larger graphs}
Table \ref{tab:merged} compares label-aware P3MP with baselines using node features and graph structure on homophilic citation graphs and larger heterophilic graphs. Classification supervision and partitions are shared; our method also uses training-label indicators in its structural encoder. We achieve the highest mean accuracy on Cora and Citeseer. On PubMed, GAT leads P3MP by 0.3 percentage points. P3MP has the highest mean accuracy on Penn94, arXiv-year, and snap-patents, with gains of 0.4, 0.9, and 1.6 percentage points over the leading baseline, respectively.

\begin{table}[t]
\caption{Label-aware P3MP and baselines that use node features and graph structure on homophilic and larger graphs. Partitions and classification supervision are shared. P3MP's structural encoder additionally receives training-label indicators. Entries are accuracy means and standard deviations over ten runs.}
\centering
\begin{adjustbox}{max width=\textwidth}
\begin{tabular}{@{}lccc|ccc@{}}
\Xhline{2\arrayrulewidth}
 & \multicolumn{3}{c}{\textbf{\underline{Homophilic}}} & \multicolumn{3}{c}{\textbf{\underline{Heterophilic}}} \\
\textbf{Dataset} 
& \textbf{Cora} & \textbf{Citeseer} & \textbf{PubMed}
& \textbf{Penn94} & \textbf{arXiv-year} & \textbf{snap-patents} \\
\rowcolor{white}
$\mathcal{G}_h$ (Eq. \ref{global_homophily})
& 0.81 & 0.74 & 0.80
& 0.47 & 0.22 & 0.07 \\
\Xhline{2\arrayrulewidth}

GCN \citep{kipf2017semi}
& 81.5$_{\pm 0.71}$ & 67.5$_{\pm 0.70}$ & 79.5$_{\pm 0.47}$
& 81.3$_{\pm 0.73}$ & 44.5$_{\pm 0.58}$ & 43.9$_{\pm 0.42}$ \\

GAT \citep{velivckovic2018graph}
& 82.6$_{\pm 0.55}$ & 68.4$_{\pm 0.83}$ & \textbf{79.9$_{\pm 0.45}$}
& 80.6$_{\pm 0.81}$ & 45.0$_{\pm 0.53}$ & 45.2$_{\pm 0.47}$ \\

H$_2$GCN \citep{zhu2020beyond}
& 80.3$_{\pm 0.52}$ & 68.5$_{\pm 0.76}$ & 78.8$_{\pm 0.37}$
& 80.4$_{\pm 0.94}$ & 47.6$_{\pm 0.41}$ & 48.1$_{\pm0.72}$ \\

GCNII \citep{chen2020simple}
& 82.2$_{\pm 0.64}$ & 67.8$_{\pm 1.21}$ & 79.4$_{\pm 0.52}$
& 81.8$_{\pm 0.63}$ & 46.1$_{\pm 0.72}$ & 47.5$_{\pm 0.60}$ \\

GPRGNN \citep{chien2021adaptive}
& 82.0$_{\pm 0.59}$ & 70.1$_{\pm 0.91}$ & 79.2$_{\pm 0.57}$
& 81.1$_{\pm 0.55}$ & 43.9$_{\pm 0.84}$ & 41.7$_{\pm 0.34}$ \\

LINKX \citep{lim2021large}
& 81.4$_{\pm 0.62}$ & 69.3$_{\pm 0.81}$ & 79.0$_{\pm 0.46}$ & 82.2$_{\pm 0.58}$ & 49.3$_{\pm 1.27}$ & 52.6$_{\pm 0.26}$ \\

\Xhline{2\arrayrulewidth}
\rowcolor{yellow!20}
\textbf{P3MP (label-aware)}
& \textbf{83.8$_{\pm 0.60}$} & \textbf{70.9$_{\pm 0.32}$} & 79.6$_{\pm 0.28}$ 
& \textbf{82.6$_{\pm 0.55}$} & \textbf{50.2$_{\pm 0.52}$} & \textbf{54.2$_{\pm 0.55}$} \\
\Xhline{2\arrayrulewidth}
\end{tabular}
\end{adjustbox}
\label{tab:merged}
\end{table}

\subsection{Prediction-time interventions}
Table \ref{tab:inference_controls} compares prediction-time interventions at selected full-model checkpoints. MAP and single-sample accuracy use ten splits on the five control graphs. The other accuracy contrasts use five checkpoints per graph. NLL uses five checkpoints on each primary graph. Paired summaries follow Appendix \ref{sec:app:controls}.

For MAP prediction, we select the modal role on each edge. For single-sample prediction, we set $K_{\mathrm{test}}=1$ at a checkpoint trained with five samples. Deterministic soft gating uses expected signed gates along its own trajectory. MC dropout averages five passes on a MAP role graph. To shuffle probabilities, we permute the complete role-probability vectors across observed edges within each graph. For graph-average replacement, we assign every edge the vector
\begin{equation}
\bar\pi=\frac1m\sum_{(i,j)\in\mathcal E_{\mathrm{obs}}}\pi_{ij}.
\end{equation}
Uniform replacement uses $(1/3,1/3,1/3)$ on every edge. We average five predictions for MC dropout, shuffling, graph-average replacement, and uniform replacement. MAP and soft gating use deterministic trajectories.

\begin{table}[ht]
\centering
\caption{Prediction-time gains with the sign convention in Eq. \ref{eq:paired-gains}. Accuracy gains are in percentage points and use five graphs with the run count shown. NLL gains use eight graphs and five checkpoints per graph, with paired 95\% intervals.}
\label{tab:inference_controls}
\begin{adjustbox}{}
\begin{tabular}{@{}lccc@{}}
\toprule
Control & Accuracy gain & Runs per graph & NLL gain \\
\midrule
Edgewise MAP roles & $+1.64$ & 10 & $+0.16\ [0.07,0.25]$ \\
Single-sample prediction & $+2.06$ & 10 & $+0.11\ [0.04,0.20]$ \\
Deterministic soft gating & $+0.94$ & 5 & $+0.08\ [0.02,0.14]$ \\
MAP with five-pass MC dropout & $+1.24$ & 5 & $+0.09\ [0.03,0.16]$ \\
Shuffled edge probabilities & $+1.34$ & 5 & $+0.11\ [0.04,0.19]$ \\
Graph-average role probabilities & $+0.82$ & 5 & $+0.10\ [0.03,0.18]$ \\
Uniform role sampling & $+3.44$ & 5 & $+0.18\ [0.08,0.28]$ \\
\bottomrule
\end{tabular}
\end{adjustbox}
\end{table}

Shuffling and graph-average replacement change different properties of the joint role law. Index the $m$ edges by $e$ and let $\sigma$ be a uniformly random permutation of these indices. Shuffling assigns $\pi_e^\sigma=\pi_{\sigma(e)}$ to edge $e$, so its permutation-averaged marginal is $\mathbb E_\sigma[\pi_e^\sigma]=\bar\pi$. Averaging over permutations gives the joint law
\begin{equation}
q_{\mathrm{shuffle}}(Z)=\frac1{m!}\sum_\sigma\prod_{e=1}^m\operatorname{Cat}(z_e;\pi_{\sigma(e)}),
\qquad
q_{\mathrm{global}}(Z)=\prod_{e=1}^m\operatorname{Cat}(z_e;\bar\pi).
\end{equation}
 For $m>1$ and distinct edges $e,f$, let
$\Sigma_{st}=m^{-1}\sum_g(\pi_g^s-\bar\pi^s)(\pi_g^t-\bar\pi^t)$ be the empirical covariance of role-probability coordinates. Sampling two distinct vectors from the permutation yields
\begin{equation}
\mathbb E_\sigma[\pi_{\sigma(e)}^s\pi_{\sigma(f)}^t]
=\bar\pi^s\bar\pi^t-\frac{\Sigma_{st}}{m-1}.
\end{equation}
For any one permutation, we reassign the probability vectors to different endpoint pairs. Averaging over permutations introduces the cross-edge dependence above. Graph-average replacement samples independently from $\bar\pi$.

\subsection{Edge specificity and annotation associations}
In Table \ref{tab:semantic_scope}, we report associations between learned role probabilities and edge annotations after model selection. We average label-agreement diagnostics over eight datasets and five evaluations per dataset. Table \ref{tab:posterior_specificity} reports the prediction change when edge-specific probabilities are replaced by their graph average.

\begin{table}[ht]
\centering
\caption{Post-hoc associations between edge probabilities and annotations across eight datasets and five evaluation runs. Diagnostics follow model selection and clean prediction.}
\label{tab:semantic_scope}
\begin{tabular}{@{}lc@{}}
\toprule
Diagnostic & Macro result \\
\midrule
Mean $\pi^+$ on same-label minus different-label edges & 0.011 \\
Mean $\pi^-$ on different-label minus same-label edges & 0.019 \\
Mean $\pi^0$ on inserted minus original edges & 0.017 \\
\bottomrule
\end{tabular}
\end{table}

Edge-specific probabilities improve the primary score over graph-average replacement on every graph in Table \ref{tab:posterior_specificity}. Each contrast averages five checkpoints. Graph-average replacement keeps the mean role vector unchanged. Uniform replacement in Table \ref{tab:inference_controls} also changes this vector.

\begin{table}[ht]
\centering
\caption{Mean paired gain from edge-specific role probabilities over graph-average replacement at the same checkpoints. Each entry averages five checkpoints using the dataset's primary percentage metric.}
\label{tab:posterior_specificity}
\begin{tabular}{@{}lcc@{}}
\toprule
Dataset & Primary metric & Mean edge-specific gain \\
\midrule
Roman-Empire & Accuracy & $+1.0$ \\
Minesweeper & ROC-AUC & $+0.7$ \\
Amazon-Ratings & Accuracy & $+1.5$ \\
Chameleon & Accuracy & $+0.8$ \\
Squirrel & Accuracy & $+0.7$ \\
Actor & Accuracy & $+0.6$ \\
Cornell & Accuracy & $+1.0$ \\
Wisconsin & Accuracy & $+0.9$ \\
\bottomrule
\end{tabular}
\end{table}

\subsection{Prediction sample count and computational cost}\label{sec:app:cost}
For Figure \ref{fig:sample_cost}, we evaluate five label-aware checkpoints per dataset at $K_{\mathrm{test}}\in\{1,2,4,5,8,16\}$. We train each checkpoint using five samples. The primary-score curve averages seven accuracies and Minesweeper ROC-AUC, and the NLL curve averages the same eight graphs.

Table \ref{tab:sample_resources} reports Chameleon prediction costs on NVIDIA A100 hardware. Let $T(K)$ be the latency of a complete prediction with $K$ samples and $M(K)$ its peak device memory. We normalize by $T(1)$ and $M(1)$ on the same graph and hardware. Latency includes shared encoding, sampling, signed propagation, classification, and probability averaging. Sequential evaluation with gradients disabled holds one sample's signed states at a time. The $K$ stored class log-probability tensors each have shape $n\times C$ and together use $4KnC$ bytes in 32-bit precision. This storage grows with sample count. At 16 samples, the measured peak memory of the complete predictor is 1.0013 times the single-sample value.

\begin{table}[ht]
\centering
\caption{Relative prediction latency and peak device memory on Chameleon using NVIDIA A100 hardware. Denominators use single-sample prediction on the same graph and hardware. Latency includes encoding, role sampling, signed propagation, classification, and probability averaging.}
\label{tab:sample_resources}
\begin{tabular}{@{}ccc@{}}
\toprule
Prediction samples & Relative latency & Relative peak memory \\
\midrule
1 & 1.00 & 1.0000 \\
2 & 1.57 & 1.0000 \\
4 & 2.78 & 1.0001 \\
5 & 3.25 & 1.0003 \\
8 & 4.96 & 1.0006 \\
16 & 8.17 & 1.0013 \\
\bottomrule
\end{tabular}
\end{table}

The shared encoders require encoding time $T_{\mathrm{shared}}$. Each sequential sample adds an approximately constant time $T_{\mathrm{sample}}$. An additive approximation to the prediction latency is
\begin{equation}
T(K)\approx T_{\mathrm{shared}}+KT_{\mathrm{sample}},\qquad
\frac{T(K)}{T(1)}\approx1+(K-1)\frac{T_{\mathrm{sample}}}{T_{\mathrm{shared}}+T_{\mathrm{sample}}}.
\end{equation}
Shared encoding runs once; signed propagation and classification run for each sample.

\subsection{Post-checkpoint graph sensitivity}
We evaluate graph and feature perturbations at checkpoints selected on clean data, keeping labels, partitions, and parameters unchanged. Edge deletion samples observed edges uniformly. Insertion adds previously unobserved edges with distinct endpoints at the same rate. The feature condition adds Gaussian noise to standardized features using a common scale across methods. Table \ref{tab:robustness_quantitative} reports clean and perturbed accuracies on Chameleon and Cornell. P3MP achieves the highest accuracy in each condition and shows smaller accuracy drops under edge deletion and insertion.

\begin{table}[!htbp]
\centering
\caption{Random-edge and feature-noise results for label-aware P3MP and baselines that use node features and graph structure. We apply each corruption to the selected predictor at evaluation time.}
\label{tab:robustness_quantitative}
\begin{adjustbox}{max width=\textwidth}
\begin{tabular}{@{}lccccc@{}}
\toprule
\textbf{Dataset} & \textbf{Method} & \textbf{Clean} & \textbf{Del. 10\%} & \textbf{Ins. 10\%} & \textbf{Feat. noise} \\
\midrule
Chameleon & GCN & 54.9 & 51.8 & 50.9 & 50.2 \\
Chameleon & GAT & 54.4 & 51.2 & 50.5 & 51.0 \\
Chameleon & TFE-GNN & 60.2 & 57.4 & 56.2 & 56.0 \\
\rowcolor{yellow!20}
Chameleon & P3MP &  62.7 & 60.9 & 60.2 & 59.4 \\
Cornell & GCN & 49.6 & 43.4 & 42.8 & 44.2 \\
Cornell & GAT & 52.4 & 46.7 & 45.8 & 45.4 \\
Cornell & DirGNN & 63.8 & 58.0 & 56.6 & 57.2 \\
\rowcolor{yellow!20}
Cornell & P3MP & 70.3 & 68.4 & 67.0 & 66.2 \\
\bottomrule
\end{tabular}
\end{adjustbox}
\end{table}

\subsection{Sensitivity under increasing corruption}\label{sec:robustness}
\begin{figure}[ht]
\centering
\includegraphics[width=\linewidth]{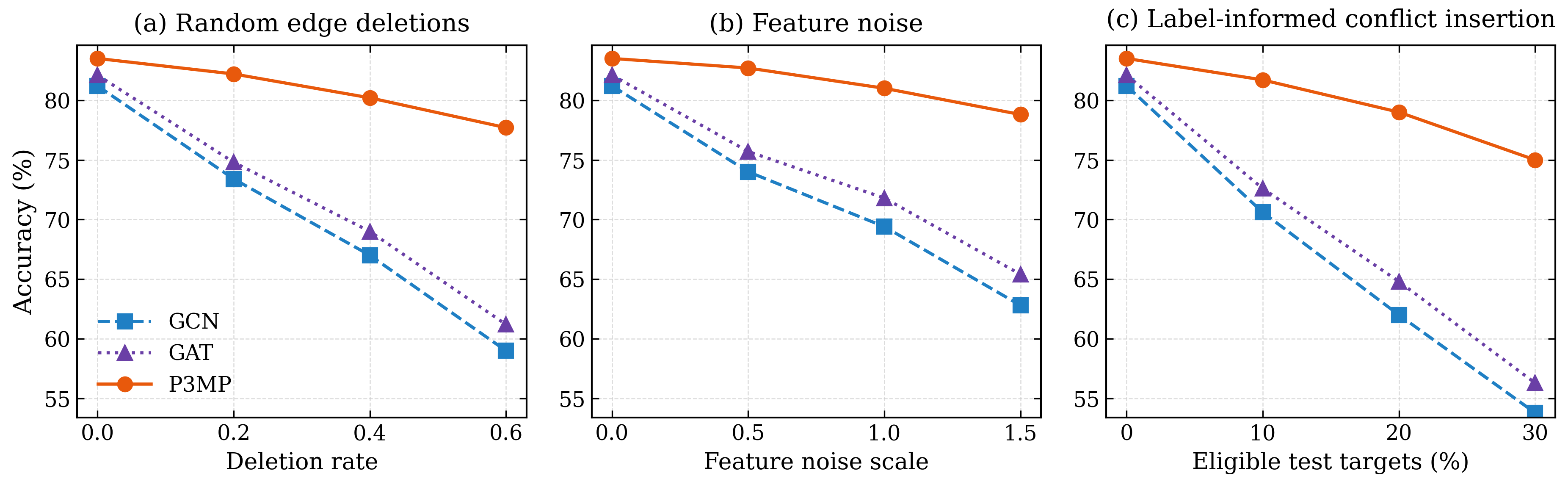}
\caption{Accuracy under increasing corruption on Cora. Panel (a) varies the random edge-deletion rate, panel (b) the feature-noise scale, and panel (c) the label-informed conflict-insertion budget from 0\% to 30\% of eligible test targets. GCN is shown with blue dashed lines and square markers, GAT with purple dotted lines and triangular markers, and P3MP with orange solid lines and circular markers.}
\label{fig:robustness}
\end{figure}

Figure \ref{fig:robustness} compares GCN, GAT, and label-aware P3MP on Cora under increasing evaluation-time corruption. Panel (a) varies random edge deletion, panel (b) varies feature-noise scale, and panel (c) shows label-informed conflict insertion. Every curve starts from the clean graph. The same perturbed graph is evaluated by each selected predictor.

\paragraph{Label-informed conflict insertion.}
For each split, we construct an ordered sequence of directed edges from standardized input features and class labels. An eligible target is a test node with a nonadjacent node from another class. Let $\mathcal C_i$ be the candidate source set of test node $i$, and let $\mathcal T_{\mathrm{elig}}$ be the eligible targets:
\begin{equation}
\begin{aligned}
\mathcal C_i&=\{j\in\mathcal V:\ j\ne i,\ A_{\mathrm{obs},ij}=0,\ y_j\ne y_i\},\\
\mathcal T_{\mathrm{elig}}&=\{i\in\mathcal V_{\mathrm{te}}:\ \mathcal C_i\ne\varnothing\}.
\end{aligned}
\end{equation}
For each eligible target, $j(i)\in\mathcal C_i$ is its closest candidate in standardized feature space, with ties resolved by the smaller node index. A permutation orders the targets as $(i_1,\ldots,i_M)$, where $M=|\mathcal T_{\mathrm{elig}}|$ is the eligible-target count. At each insertion level, we use a prefix of this sequence.

\paragraph{Percentage budget.}
We express the insertion budget as a percentage of eligible test targets. For a requested percentage $\beta\in[0,100]$, the number of inserted directed edges is
\begin{equation}
 B(\beta)=\left\lfloor\frac{\beta M}{100}\right\rfloor,\qquad
 \mathcal E^{(\beta)}=\mathcal E_{\mathrm{obs}}\cup
 \{(i_t,j(i_t)):1\le t\le B(\beta)\}.
\end{equation}
Each selected target receives one incoming edge, so $B(\beta)$ counts both added edges and perturbed targets. For $M=0$, the graph stays unchanged. All inserted pairs are distinct, and a source may be assigned to several targets. Algorithm \ref{alg:conflict_insertion} specifies the construction.

\begin{algorithm}[ht]
\caption{Label-informed conflict insertion with a percentage budget}\label{alg:conflict_insertion}
\begin{algorithmic}[1]
\REQUIRE Clean graph $\mathcal G_{\mathrm{obs}}$, standardized input features, labels $y$, test set $\mathcal V_{\mathrm{te}}$, target percentage $\beta\in[0,100]$
\STATE Form $\mathcal C_i$ for every test node and collect $\mathcal T_{\mathrm{elig}}$
\STATE Set $M\gets|\mathcal T_{\mathrm{elig}}|$ and $B\gets\lfloor\beta M/100\rfloor$
\STATE Order the eligible targets as $(i_1,\ldots,i_M)$ using a permutation
\FOR{each eligible target $i$ in that order}
\STATE Select the closest source $j(i)\in\mathcal C_i$ in standardized feature space, resolving ties by node index
\STATE Append $(i,j(i))$ to the ordered insertion list
\ENDFOR
\STATE Add the first $B$ pairs to $\mathcal E_{\mathrm{obs}}$ to obtain $\mathcal E^{(\beta)}$
\STATE Evaluate every selected predictor on $(\mathcal V,\mathcal E^{(\beta)})$ with unchanged parameters
\end{algorithmic}
\end{algorithm}

In Figure \ref{fig:robustness}(c), we evaluate $\beta \in \{0,10,20,30\}$, corresponding to insertion budgets from 0\% to 30\% of eligible test targets.

After checkpoint selection on the clean graph, class labels determine disagreement, and feature distances select the inserted sources. Test labels enter this diagnostic construction and the accuracy calculation. The perturbations are label-informed and model-independent. Accuracy decreases as corruption grows. Across the plotted corruption levels, P3MP has higher accuracy and a smaller reduction from its clean-graph accuracy. Table \ref{tab:robustness_quantitative} reports the complementary random-edge and feature-noise results on Chameleon and Cornell.

\newpage
\section{Limitations}
We derive the two-layer preactivation decomposition for coupled computations with common parameters. Retraining also changes the learned parameters and regularization. The finite-margin identity describes the loss change, whose sign depends on the readout and margin responses. The activity-order construction uses one sampled self-edge with zero self-maps. P3MP assumes conditionally independent roles, and we evaluate it in transductive settings. Inductive generalization and correlated role sampling remain extensions of this setting. In Appendix \ref{sec:robustness}, we evaluate empirical sensitivity at increasing perturbation levels. Certified worst-case robustness requires a separate theoretical analysis.

\end{document}